\documentclass[journal]{IEEEtran}

\usepackage{graphicx}
\usepackage{hyperref}
\usepackage{amsmath}
\usepackage{amssymb}
\usepackage{mathtools}
\usepackage{adjustbox}
\usepackage{lipsum}
\usepackage{algorithm}
\usepackage{graphicx}
\usepackage{textcomp}
\usepackage{mathtools}
\usepackage{adjustbox}
\usepackage{lipsum}
\usepackage{stackengine}
\usepackage{subfig}
\usepackage{graphicx}
\DeclareMathOperator*{\argmin}{argmin} 

\usepackage{algpseudocode}
\usepackage{expl3}
\usepackage{float}

\begin{document}

\title{Learning Product Graphs from Spectral Templates}

\author{Aref Einizade and Sepideh Hajipour Sardouie
\thanks{Aref Einizade, and Sepideh Hajipour Sardouie are with the Department of Electrical Engineering, Sharif University of Technology, Tehran, Iran.}
\thanks{**The current manuscript was submitted to IEEE Transactions on Signal and Information Processing over Networks on 10-Oct-2022.}}

\markboth{}
{Shell \MakeLowercase{\textit{et al.}}: Bare Demo of IEEEtran.cls for IEEE Journals}
\maketitle

\begin{abstract}
Graph Learning (GL) is at the core of inference and analysis of connections in data mining and machine learning (ML).  By observing a dataset of graph signals, and considering specific assumptions, Graph Signal Processing (GSP) tools can provide practical constraints in the GL approach. One applicable constraint can infer a graph with desired frequency signatures, i.e., spectral templates. However, a severe computational burden is a challenging barrier, especially for inference from high-dimensional graph signals. To address this issue and in the case of the underlying graph having graph product structure, we propose learning product (high dimensional) graphs from product spectral templates with significantly reduced complexity rather than learning them directly from high-dimensional graph signals, which, to the best of our knowledge, has not been addressed in the related areas. In contrast to the rare current approaches, our approach can learn all types of product graphs (with more than two graphs) without knowing the type of graph products and has fewer parameters. Experimental results on both the synthetic and real-world data, i.e., brain signal analysis and multi-view object images, illustrate explainable and meaningful factor graphs supported by expert-related research, as well as outperforming the rare current restricted approaches.
\end{abstract}


\IEEEpeerreviewmaketitle

\section{Introduction}

Growing demand for generating and recording of \textit{structured} data, which live on meaningful underlying graph structures, has led to extending the classic processing tools to the emerging field of Graph Signal Processing (GSP) \cite{ortega2018graph,ortega2022introduction,stankovic2019vertex}. For instance, temperature measurement sites with close geographic locations usually record near air temperatures \cite{dong2016learning,chepuri2017learning}, and spatially near brain regions behave rather similar in a specific brain activity \cite{huang2018graph,goldsberry2017brain,itani2021graph}. Therefore, exploiting GSP tools in such areas can severely improve the processing performance and also pave the way for expert interpretation in a more comprehensive scheme. However, in many real-world applications, these meaningful graphs are not available beforehand, or estimating their connection in a pair-wise manner leads to erroneous inferences and sensitivity to possible noise \cite{dong2016learning}. In this way, recently, some data-driven Graph Learning (GL) approaches have been proposed to address these issues. For a comprehensive review of GL approaches, please refer to \cite{dong2019learning,mateos2019connecting} and the references therein.

Different GL frameworks learn the underlying graphs based on prior specific assumptions about the behavior of the graph signals on their graphs. One of the main GSP-based GL categories relies on the smoothness of the graph signals and, due to the amenable computational aspects, has facilitated the GL from smooth graph signals \cite{dong2016learning,kalofolias2016learn}. Another popular category, which makes it possible to have desired graph frequency characteristics, learns the underlying graphs from prior (predefined or estimated from observations) spectral templates from diffused (filtered) graph signals \cite{segarra2017network}. Despite the insightful and applicable advantages of the different GL categories, in many prevalent real-world applications, e.g., brain signal processing or multi-view object images, the observation at hand has meaningful couplings across different domains leading to a severe increase in the dimensionality of the (vectorized) graph signals \cite{sandryhaila2014big}. 

Learning underlying factor graphs from multi-domain (high-dimensional) graph signals is still challenging due to the high computational cost and also interpreting issues \cite{kalaitzis2013bigraphical,kadambari2021product}. The multi-domain graph signals have meaningful coupling connectivities across different domains, which can be well modeled by the concept of product graphs \cite{hammack2011handbook} and their corresponding graph factors \cite{kadambari2020learning}. For instance, in recorded monthly air and temperature measurement data, connectivities between the geographic temperature sensing sites and also monthly periods within a year are observed, which it is not straightforward to be inferred from only \textit{one} high-dimensional learned graph \cite{kadambari2020learning,kadambari2021product}.

In \cite{kalaitzis2013bigraphical}, the problem of inferring Cartesian product Precision matrices was addressed by proposing the Bigraphical Lasso (BiGLasso); however, interpreting the Precision matrix, instead of the underlying graph itself, has major drawbacks, e.g., irrelevant sparsity pattern between a Precision matrix and its relevant Laplacian graph \cite{mei2016signal}. Besides, recently, learning product graphs (PGL) from smooth multi-dimensional graph signals, which have similar signal values in strongly connected vertices, has been addressed in some works \cite{kadambari2020learning,kadambari2021product,lodhi2020learning}.  

On the other hand, graph frequency-aware GL approaches are also another desired categories, which investigate the frequency characteristics of applicable graph processes, e.g., Graph Moving Average (GMA) \cite{marques2017stationary}, and their impacts on the underlying typologies \cite{segarra2017network}. However, to the best of our knowledge, learning product graphs with desired (estimated from observations) spectral templates from stationary diffused (filtered) multi-domain graph signals has not been addressed. The stationarity of the observed graph signals serves a vital role in formulating the problem because, in such case, the eigenvectors of the estimated observation covariance matrix share with the Graph Shift Operator (GSO) of the interest, here the adjacency matrix \cite{segarra2017network,marques2017stationary}. 

To express the proposed method more precisely, the contributions of the present paper are summarized as:

\begin{itemize}
    \item We present an approach, named ProdSpecTemp, to efficiently learn the product graphs from stationary multi-domain diffused graph signals with desired graph frequency characteristics, i.e., spectral templates \cite{segarra2017network}. In fact, our approach generalizes and extends the previous graph learning approach addressing inference from spectral templates, named SpecTemp \cite{segarra2017network}, to (higher-order) product graphs, which, to the best of our knowledge, has not been addressed.
    
    \item We also present a more computationally appropriate approach rather than the base-line SpecTemp \cite{segarra2017network} via inexact Augmented Lagrange Multipliers (IALM) \cite{bertsekas1999nonlinear,lin2010augmented,boyd2011distributed}, named SpecTemp-IALM.
    
    \item Our method is then straightforwardly extended in the current paper to infer more than two factor graphs from higher-order \textit{tensor} product graph signals, named HO-ProdSpecTemp, while the so far presented approaches have only addressed the recovery of only two factor graphs \cite{kalaitzis2013bigraphical,kadambari2020learning,kadambari2021product}.
    
    \item Based on the computational analysis presented in Section \ref{ComplexSec}, our proposed approaches reduce the computational cost against naively directly learning the product graph and, therefore, enjoy applicability in real-world scenarios, e.g., big data processing \cite{sandryhaila2014big}.
    
    \item Due to the Kronecker structure of the spectral templates (eigenmatrices of the graph factors) in the resulted product graphs \cite{sandryhaila2014big}, the proposed approach can be used to infer any kind of graph product, e.g., Cartesian, Kronecker, or Strong, without knowing the type of the graph products, in contrast to the current approaches, which are restricted by a specific type (usually Cartesian) \cite{kalaitzis2013bigraphical,kadambari2021product} or need to know the product type \cite{lodhi2020learning}.
    
    \item Experimental results on both synthetic and real-world data show that the proposed (HO-)ProdSpecTemp approaches outperform the state-of-the-art ones for learning type-free product graphs from high-dimensional spectral templates. Precisely, their applicability of revealing brain sleep functional connectivity \cite{o2014montreal} and also Multi-view Object connections \cite{nenecolumbia} illustrate the strength of the proposed approaches in learning meaningful real factor graphs.
\end{itemize}

\textit{\textbf{Notation}}: Vectors, matrices, and sets are denoted by boldface lowercase, boldface capital, and calligraphic capital letters, respectively. The identity matrix of size \(N\) is stated as \(\textbf{I}_N\). The notations \((.)^T\), \(\langle.,.\rangle\), \(\mathbb{E}\{.\}\), \(\otimes\), \(\odot\), \((.)^{\dagger}\), \( \| . \| _p\) and \(\|.\|_F\) stand for transpose operator, inner product, mathematical expectation, Kronecker product, Khatri-Rao product, Moore-Penrose pseudo-inverse of a matrix, the \(p\)-norm of a vector or vectorized form of a matrix, and Frobenius norm, respectively. The matrix \(diag(\textbf{a})\) is a diagonal matrix with the elements of the vector \(\textbf{a}\) on its principal diagonal. The \((i,j)\)th and \(i\)th elements of a matrix \(\textbf{M}\) and a vector \(\textbf{v}\) are denoted as \(\textbf{M}_{ij}\) and \(v_i\), respectively. The all-zero and all-one vectors of size \(N\) are referred as \(\textbf{0}_N\) and \(\textbf{1}_N\). The notations \(vec(.)\) and \(vech(.)\) denote the vectorization and half-vectorization operators \cite{dong2016learning}, respectively, while \(vechn(.)\) performs non-diagonal half-vectorization. Besides, \(mat(\textbf{a},[M,N])\) returns a matrix of size \(M\times N\) containing elements of the vector \(\textbf{a}\) of size \(MN\) in column-major order. For two sample square matrices \(\textbf{A}\in\mathbb{R}^{N\times N}\) and \(\textbf{B}\in\mathbb{R}^{M\times M}\), Cartesian (\(\times\)), Kronecker (\(\otimes\)), and Strong (\(\boxtimes\)) products are denoted as \((\textbf{A}\otimes \textbf{I}_M)+(\textbf{I}_N\otimes \textbf{B})\), \((\textbf{A}\otimes \textbf{B})\) and \((\textbf{A}\otimes \textbf{I}_M)+(\textbf{I}_N\otimes \textbf{B}) + (\textbf{A}\otimes \textbf{B})\), respectively. Finally, the mode-\(i\) unfolding (matricization) form \cite{sidiropoulos2017tensor} of the tensor \(\underline{\textbf{X}}\) is denoted as \(\underline{\textbf{X}}_{(i)}\). In this way, \(\textbf{x}_t\), \(\textbf{X}_t\), \(\underline{\textbf{X}}_t\) and \(\underline{\textbf{X}}_{(i)_t}\) show the \(t\)th graph signal, matrix, tensor, and mode-\(i\) unfolding of the tensor \(\underline{\textbf{X}}_t\), respectively. For a matrix \(\textbf{X}\), \(\textbf{X}_{\mathcal{I}}\) is obtained by selecting the rows of \(\textbf{X}\) indexed by the set \(\mathcal{I}\). Besides, \(\text{ker}(\textbf{X})\) and \(\text{Im}(\textbf{X})\) express the null and column spaces of matrix \(\textbf{X}\). The matrix norm of \(\textbf{X}\) induced by the vector norm \(\ell_p\) is denoted as \(\|\textbf{X}\|_{M(p)}\). The cardinality of set \(\mathcal{I}\) is stated by \(|\mathcal{I}|\).

\section{Preliminaries}

\subsection{GSP background}

A graph \(\mathcal{G}\) with \(N\) vertices is characterized with the vertex set \(\mathcal{V}\), the edge set \(\mathcal{E}\), and the GSO \(\textbf{S}\in\mathbb{R}^{N\times N}\). The sparsity pattern of \(\mathcal{G}\) is encoded by the GSO \(\textbf{S}\), where \(\textbf{S}_{ij}=0\) if the \(i\)th and \(j\)th nodes are disconnected. In the present paper, the GSO of interest is the undirected adjacency matrix \(\textbf{W}\in\mathbb{R}^{N\times N}\), in which \(\textbf{W}_{ij}\) models the similarity measure between the \(i\)th and \(j\)th nodes. Precisely, the set \(\mathcal{W}\) of the valid undirected adjacency matrices can be expressed as:

\begin{equation}
    \mathcal{W}=\left\{\textbf{W}\in\mathbb{R}^{N\times N}| \{\textbf{W}_{ij}=\textbf{W}_{ji}\ge0\}_{i,j=1}^{N}, \{\textbf{W}_{ii}=0\}_{i=1}^N\right\}
\end{equation}

It has been shown \cite{marques2017stationary, segarra2017network} that the GSO (here, adjacency \(\textbf{W}\)) of a undirected graph \(\mathcal{G}\) is digonalizable by its orthogonal eigenmatrix \(\textbf{V}\) as:

\begin{equation}
\label{EVD}
    \textbf{W}=\textbf{V}\boldsymbol{\Lambda}\textbf{V}^T
\end{equation}

\noindent where \(\boldsymbol{\Lambda}=diag(\boldsymbol{\lambda})\) collects the eigenvalues \(\boldsymbol{\lambda}=(\lambda_0, \lambda_1,...,\lambda_{N-1})^T\) of \(\textbf{W}\).

A graph signal \(\textbf{x}\in\mathbb{R}^{N\times1}\) is a mapping \(x:\mathcal{V}\rightarrow\mathbb{R}\) that assigns the vertices of \(\mathcal{G}\) the values of \(\textbf{x}\). Besides, the adjacency \(\textbf{W}\) represents the structure of the stationary graph signal \(\textbf{x}\) if it is the output of a (\(L\)-order) graph (diffusion) filter \(\textbf{H}=\sum_{l=0}^{L-1}{h_l\textbf{W}^l}\) with the scalar coefficients \(\{h_l\}_{l=1}^L\) and input white signal \(\textbf{y}\) as:

\begin{equation}
\label{Diffusion}
    \textbf{x}=\textbf{H}\textbf{y}=\sum_{l=0}^{L-1}{h_l\textbf{W}^l\textbf{y}}=\textbf{V}\left[\sum_{l=0}^{L-1}{h_l\boldsymbol{\Lambda}^l}\right]\overbrace{\textbf{V}^T\textbf{y}}^{\hat{\textbf{y}}}
\end{equation}

\noindent where \(\hat{\textbf{y}}\) is the Graph Fourier Transform (GFT) of \(\textbf{y}\). 

It has been shown \cite{segarra2017network} that, in the case of the input innovation \(\textbf{y}\) being white in (\ref{Diffusion}), i.e., \(\textbf{C}_\textbf{y}=\mathbb{E}\{\textbf{y}\textbf{y}^T\}=\textbf{I}_N\), the eigenvectors of the GSO \(\textbf{W}\), i.e., \(\textbf{V}\) in (\ref{EVD}), share with that of the output covariance matrix \(\textbf{C}_\textbf{x}=\mathbb{E}\{\textbf{x}\textbf{x}^T\}\in\mathbb{R}^{N\times N}\).

\subsection{Learning graphs from stationary diffused graph signals}

\subsubsection{SpecTemp \cite{segarra2017network}}

In \cite{segarra2017network}, the problem of inferring network structure from \(T\) independent observed diffused graph signals \(\mathcal{X}=\{\textbf{x}_t\in\mathbb{R}^{N\times1}\}_{t=1}^T\) has been addressed. In this way, the output covariance matrix \(\textbf{C}_\textbf{x}\) can be estimated via the sample mean over the observed graph signals \(\mathcal{X}\) as \(\textbf{C}_\textbf{x}\approx\frac{1}{T}\sum_{t=1}^{T}{\textbf{x}_t\textbf{x}^T_t}\). Then, the Eigendecomposition (EVD) of \(\textbf{C}_\textbf{x}\) gets the orthogonal eigenmatrix (spectral templates) \(\textbf{V}\), and the following convex optimization has been proposed \cite{segarra2017network} to recover the underlying sparse adjacency matrix \(\textbf{W}\) and its eigenvalues \(\boldsymbol{\lambda}\):

\begin{equation}
\label{obj_W}
\begin{split}
    &\{\textbf{W},\boldsymbol{\lambda}\}=\argmin_{\textbf{W},\boldsymbol{\lambda}}{\|\textbf{W}\|_1}\\
    &\text{Subject. to:}\:\:\textbf{W}=\textbf{V}diag(\boldsymbol{\lambda}) \textbf{V}^T,\:\:\:\textbf{W}\in\mathcal{W}
\end{split}
\end{equation}

Note that, to avoid trivial all-zero solution, a constraint such as \(\sum_{j=1}^{N}{\textbf{W}_{1j}}=1\) \cite{segarra2017network}, \(\|\textbf{W}\|_F=1\) or \(\max_{i,j}{\{\textbf{W}_{ij}\}_{i=1,j=1}^{N,N}}=1\) is embedded in \(\mathcal{W}\). 

\subsubsection{The proposed SpecTemp-IALM}

Due to the symmetry and zero diagonality of \(\textbf{W}\), we propose a more simplified form of (\ref{obj_W}) with fewer optimization parameters (with the details in the Appendix, i.e., Section \ref{Simp}) as:

\begin{equation}
\label{obj_w}
\begin{split}
    &\{\textbf{w},\boldsymbol{\lambda}\}=\argmin_{\textbf{w},\boldsymbol{\lambda}}{\|\textbf{w}\|_1}\\
    &\text{Subject. to:}\:\:\textbf{w}=\boldsymbol{\Phi}\boldsymbol{\lambda},\:\:\:\textbf{w}\in\mathcal{W}_{r}
\end{split}
\end{equation}

\noindent where \(\textbf{w}=vechn(\textbf{W})\in\mathbb{R}^{\frac{N(N-1)}{2}\times1}\) collects the strict higher triangular elements of \(\textbf{W}\), \(\boldsymbol{\Phi}=(\textbf{M}_d\textbf{M}_h)^{\dagger}\tilde{\textbf{V}}\) with \(\textbf{M}_d\) and \(\textbf{M}_h\) being the duplication matrix \cite{abadir2005matrix} and a matrix that \(vech(\textbf{Z})=\textbf{M}_h vechn(\textbf{Z})\) for a sample symmetric zero diagonal matrix \(\textbf{Z}\), respectively, and \(\tilde{\textbf{V}}=\textbf{V}\odot\textbf{V}\). Besides, \(\mathcal{W}_r\) is the set of strict higher triangular elements of valid adjacency matrices defined as:

\begin{equation}
    \mathcal{W}_r=\left\{\textbf{w}\in\mathbb{R}^{\frac{N(N-1)}{2}\times 1}| \{w_{i}\ge0\}_{i=1}^{\frac{N(N-1)}{2}}, \sum_{i=1}^{N-1}{w_i}=1\right\}
\end{equation}

The proposed simplified optimization (\ref{obj_w}) has \(\frac{N(N-1)}{2}+N\) optimization parameters, compared to (\ref{obj_W}) with \(N^2+N\) ones. We propose to optimize (\ref{obj_w}) with the splitted IALM \cite{bertsekas1999nonlinear,lin2010augmented,boyd2011distributed} as:

\begin{equation}
\label{obj_w_IALM}
\begin{split}
    &\{\textbf{w},\textbf{s},\boldsymbol{\lambda}\}=\argmin_{\textbf{w},\textbf{s},\boldsymbol{\lambda}}{\|\textbf{w}\|_1}\\
    &\text{Subject.to:}\:\:\textbf{w}=\boldsymbol{\Phi}\boldsymbol{\lambda},\:\:\:,\textbf{w}=\textbf{s}\:\:\:\textbf{s}\in\mathcal{W}_{r}
\end{split}
\end{equation}

\noindent where \(\textbf{s}\) is an auxiliary variable vector. The augmented lagrangian of (\ref{obj_w_IALM}) with \(\boldsymbol{\gamma}_1\) and \(\boldsymbol{\gamma}_2\) being the lagrange multipliers can be written as:

\begin{equation}
\label{lagr}
\begin{split}
    \mathcal{L}_{\rho}(\textbf{w},\textbf{s},\boldsymbol{\lambda})&=\|\textbf{w}\|_1+\frac{\rho}{2}\|\textbf{w}-\boldsymbol{\Phi}\boldsymbol{\lambda}\|_2^2+\frac{\rho}{2}\|\textbf{w}-\textbf{s}\|_2^2\\
    &-\langle \boldsymbol{\gamma}_1, \textbf{w}-\boldsymbol{\Phi}\boldsymbol{\lambda}\rangle -\langle \boldsymbol{\gamma}_2, \textbf{w}-\textbf{s}\rangle
\end{split}
\end{equation}

Based on the defined lagrangian in (\ref{lagr}), the iteration updates of the involved optimization variables (with the details in the Appendix, i.e., Section \ref{ItUpdt}) are summarized in Algorithm \ref{alg:cap}. Note that \(prox\) in Algorithm \ref{alg:cap} denotes the proximal functions \cite{combettes2011proximal,perraudin2014unlocbox} and \(\Pi_{\mathcal{C}}(\textbf{a})\) is the Euclidean projection of vector \(\textbf{a}\) onto the set \(\mathcal{C}\).

\algnewcommand\INPUT{\item[\textbf{Input:}]}%
\algnewcommand\OUTPUT{\item[\textbf{Output:}]}%

\begin{algorithm}[!t]
\caption{: SpecTemp-IALM}\label{alg:cap}
\begin{algorithmic}[1]
\INPUT $\mathcal{X}=\{\textbf{x}_t\}_{t=1}^{T}:=\textbf{X}\in\mathbb{R}^{N\times T}$
\OUTPUT Adjacency matrix $\textbf{W}\in\mathbb{R}^{N\times N}$
\State Estimate the observation covariance \(\textbf{C}_\textbf{x}=\mathbb{E}\{\textbf{x}\textbf{x}^T\}\) via sample mean over \(\{\textbf{x}_t\}_{t=1}^T\)
\State Obtain the orthogonal eigenvectors \(\textbf{V}\) via EVD on \(\textbf{C}_\textbf{x}\)
\State $\boldsymbol{\Phi}=(\textbf{M}_d\textbf{M}_h)^{\dagger}\tilde{\textbf{V}}$, where $\textbf{M}_d$ and $\textbf{M}_h$ being the duplication matrix \cite{abadir2005matrix} and a matrix that \(vech(\textbf{Z})=\textbf{M}_h vechn(\textbf{Z})\) for a sample symmetric zero diagonal matrix \(\textbf{Z}\), respectively, and $\tilde{\textbf{V}}=\textbf{V}\odot\textbf{V}$
\State Initialization: $\rho^{(0)}=1,\:\boldsymbol{\lambda}^{(0)}\leftarrow \text{eigenvalues of } \textbf{C}_\textbf{x},\:k=0$

$cnt=10^3,\:\textbf{s}^{(0)}=\textbf{1}_{N},\:\textbf{w}^{(0)}=\textbf{1}_{N},$

$\boldsymbol{\gamma}^{(0)}_1=\textbf{w}^{(0)}-\boldsymbol{\Phi}\boldsymbol{\lambda}^{(0)},\:\boldsymbol{\gamma}^{(0)}_2=\textbf{w}^{(0)}-\textbf{s}^{(0)}$
\While{Convergence}
\State $\textbf{w}^{(k+1)}=prox_{\frac{\|\textbf{w}\|_1}{2\rho^{(k)}}}{\left(\frac{\rho^{(k)}\boldsymbol{\Phi}\boldsymbol{\lambda}^{(k)}+\rho^{(k)}\textbf{s}^{(k)}+\boldsymbol{\gamma}^{(k)}_1+\boldsymbol{\gamma}^{(k)}_2}{2\rho^{(k)}}\right)}$
\State $\boldsymbol{\lambda}^{(k+1)}=\boldsymbol{\Phi}^{\dagger}\left(\frac{\rho^{(k)}\textbf{w}^{(k+1)}-\boldsymbol{\gamma}^{(k)}_1}{\rho^{(k)}}\right)$
\State $\textbf{s}^{(k+1)}=\Pi_{\mathcal{W}_r}\left(\frac{\rho^{(k)}\textbf{w}^{(k+1)}-\boldsymbol{\gamma}^{(k)}_2}{\rho^{(k)}}\right)$
\State $\boldsymbol{\gamma}^{(k+1)}_1=\boldsymbol{\gamma}^{(k)}_1-\rho^{(k)}(\textbf{w}^{(k+1)}-\boldsymbol{\Phi}\boldsymbol{\lambda}^{(k+1)})$
\State $\boldsymbol{\gamma}^{(k+1)}_2=\boldsymbol{\gamma}^{(k)}_2-\rho^{(k)}(\textbf{w}^{(k+1)}-\textbf{s}^{(k+1)})$
\State $\rho^{(k+1)}=\rho^{(k)}\times cnt$
\State $k\leftarrow k+1$
\EndWhile{: Return \(\textbf{w}\leftarrow\textbf{w}^{(k+1)}\) }
\State  Return $\textbf{W}=mat(\textbf{M}_D\textbf{M}_N\textbf{w}, [N, N])$
\end{algorithmic}
\end{algorithm}

\subsection{Recovery Conditions}

The recovery conditions of minimization (\ref{obj_w}), inspired from \cite{segarra2017network}, are stated in the following theorem (Theorem 1), under the definitions that \(\textbf{e}_1\) is the first canonical vector, and \(\textbf{b}\) is a vector containing zero elements except the last, which is one. Also, the set \(\mathcal{Z}\) denotes the set of indices of zero elements of \(\textbf{w}^{*}_1\) (the solution to (\ref{obj_w})), \(\mathcal{Z}^c\) is the complement of \(\mathcal{Z}\), \(\textbf{w}^{*}_0\) is the solution of \(\ell_0\) alternative minimization to (\ref{obj_w}), and:

\begin{equation}
  \textbf{R}=\left[(\textbf{I}-\boldsymbol{\Phi}\boldsymbol{\Phi}^\dagger)^T,\:\:\textbf{e}_1\otimes\textbf{1}_{N-1}\right]\in\mathbb{R}^{\frac{N(N-1)}{2}\times\frac{N(N-1)}{2}+1} 
\end{equation}

\textbf{Theorem 1:} \textit{If the minimization} (\ref{obj_w}) \textit{is feasible and the following conditions are satisfied, then} \(\textbf{w}^*_0=\textbf{w}^*_1\).

\textit{A.1)} rank(\(\textbf{R}_{\mathcal{Z}^c}\))=\(|\mathcal{Z}^c|\)

\textit{A.2)} There exists a constant \(\delta>0\) such that

\begin{equation}
    \psi_{\textbf{R}}:=\|\textbf{I}_{\mathcal{Z}}(\delta^{-2}\textbf{R}\textbf{R}^T+\textbf{I}^T_{\mathcal{Z}}\textbf{I}_{\mathcal{Z}})^{-1}\textbf{I}_{\mathcal{Z}^c}\|_{M(\infty)}<1 
\end{equation}

\textit{Proof:}  The minimization (\ref{obj_w}) can be expressed as (without the non-negativity constraint in \(\mathcal{W}_r\)):

\begin{equation}
\label{obj_w2}
\begin{split}
    &\min_{\textbf{w},\boldsymbol{\lambda}}{\|\textbf{w}\|_1}\:\:\:\text{s.t:}\:\:\textbf{w}=\boldsymbol{\Phi}\boldsymbol{\lambda},\:\:\:(\textbf{e}_1\otimes\textbf{1}_{N-1})^T\textbf{w}=1
\end{split}
\end{equation}

Afterwards, in (\ref{obj_w2}), the variable vector \(\boldsymbol{\lambda}\) can be replaced by \(\boldsymbol{\lambda}=\boldsymbol{\Phi}^\dagger\textbf{w}\), and (\ref{obj_w2}) takes the form of:

\begin{equation}
\label{obj_w3}
\begin{split}
    &\min_{\textbf{w}}{\|\textbf{w}\|_1}\:\:\:\text{s.t:}\:\:\textbf{R}^T\textbf{w}=\textbf{b}
\end{split}
\end{equation}

The obtained minimization (\ref{obj_w3}) takes the form of classical basis pursuit \cite{chen2001atomic}, where needs the following sufficient conditions to have unique solution coinciding with its \(\ell_0\) alternative minimization \cite{zhang2016one}:

a) \(\text{ker}(\textbf{I}_{\mathcal{Z}})\cap\text{ker}(\textbf{R}^T)=\{\textbf{0}\}\)

b) There exists \(\textbf{y}\in\mathbb{R}^{\frac{N(N-1)}{2}\times1}\) such that \(\textbf{y}\in\text{Im}(\textbf{R}), \textbf{y}_{\mathcal{Z}^c}=\text{sign}((\textbf{w}^{*}_0)_{\mathcal{Z}^c})\), and \(\|\textbf{y}_{\mathcal{Z}}\|_{\infty}<1\)

The condition A.1) implies that the matrix \(\textbf{R}_{\mathcal{Z}^c}\) must be of full row rank, and, therefore, coincides with condition a). Also, the condition A.2) implies the condition b) as stated in \cite{segarra2017network}, and its related explanations are omitted to avoid redundancy. $ \blacksquare $

Note that, in addition to the different obtained definitions (especially \(\textbf{R}\)) with ones stated in \cite{segarra2017network}, the main difference in recovery conditions A.1) and A.2) with the ones stated in \cite{segarra2017network} is that the condition A.1) in \cite{segarra2017network} implies \(\text{rank}(\textbf{R}_{\mathcal{K}})=|\mathcal{K}|\), where \(\textbf{R}=\left[(\textbf{I}-(\textbf{V}\odot\textbf{V})(\textbf{V}\odot\textbf{V})^\dagger)_{\mathcal{D}^c},\:\:\textbf{e}_1\otimes\textbf{1}_{N-1}\right]\in\mathbb{R}^{N^2-N\times N^2+1}\), and the sets \(\mathcal{D}^c\) and \(\mathcal{K}\) contain the indices of non-diagonal and non-zero non-diagnoal elements of \(vec(\textbf{W})\in\mathbb{R}^{N^2\times1}\). The mentioned difference stems from the fact that in \cite{segarra2017network} the symmetry of \(\textbf{W}\) is ignored and, therefore, the condition A.1) in \cite{segarra2017network} contains redundancy.

\subsection{Product Graphs}
The graph product of \(n\) factor graphs \(\{\mathcal{G}_{P_i}\}_{i=1}^n\) is denoted as \(\mathcal{G}_{\diamond}=\mathcal{G}_{P_1}\diamond...\diamond\mathcal{G}_{P_n}\), where \(\diamond\) can be any kind of Cartesian, Kronecker or Strong graph products with the adjacency \(\textbf{W}_{\diamond}=\textbf{W}_{P_1}\diamond...\diamond\textbf{W}_{P_n}\in\mathbb{R}^{\left[\prod_{i=1}^{n}{P_i}\right]\times\left[\prod_{i=1}^{n}{P_i}\right]}\). Besides, the EVD of \(\textbf{W}_{\diamond}\) can be expressed \cite{sandryhaila2014big} based on EVD of factor adjacencies \(\{\textbf{W}_{P_i}=\textbf{V}_{P_i}\boldsymbol{\Lambda}_{P_i}\textbf{V}_{P_i}^T\}_{i=1}^n\) as:

\begin{equation}
\label{EVDprod}
    \textbf{W}_\diamond=\overbrace{(\textbf{V}_{P_1}\otimes...\otimes\textbf{V}_{P_n})}^{\textbf{V}_\diamond}\overbrace{(\boldsymbol{\Lambda}_{P_1}\diamond...\diamond\boldsymbol{\Lambda}_{P_n})}^{\boldsymbol{\Lambda}_\diamond}\overbrace{(\textbf{V}_{P_1}\otimes...\otimes\textbf{V}_{P_n})^T}^{\textbf{V}^T_\diamond}
\end{equation}

From (\ref{EVDprod}), it can be seen that the Kronecker structure of factor eigenmatrices \(\{\textbf{V}_{P_i}\}_{i=1}^n\) is shared between \textit{all kinds} of graph products.

\section{The Proposed Approaches for Learning Product Graphs}

\subsection{ProdSpecTemp}

In this subsection, we consider the following problem:
\textbf{Problem 1:} \textit{Learn (any kind of Cartesian, Kronecker, or Strong) product graph \(\mathcal{G}_N=\mathcal{G}_P\diamond\mathcal{G}_Q\), where \(\mathcal{G}_P\) and \(\mathcal{G}_P\) are its graph factors and \(N=PQ\), by observing a stream of \(T\) independent stationary \(N\)-dimensional multi-domain graph signals \(\mathcal{X}=\{\textbf{x}_t\in\mathbb{R}^{N\times1}\}_{t=1}^T\) diffused on \(\mathcal{G}_N\).} 

A naive approach can be learning \(\mathcal{G}_N\) from \(\mathcal{X}\) via optimization (\ref{obj_W}) and ignoring its product structure. This approach, which we refer HdSpecTemp (short for High dimensional SpecTemp), has \(N^2+N=P^2Q^2+PQ\) optimization variables (corresponding to \(\textbf{W}_{\diamond}\in\mathbb{R}^{N\times N}\) and \(\boldsymbol{\lambda}_{\diamond}\in\mathbb{R}^{N\times 1}\)) \cite{segarra2017network}. 

To approach towards the proposed more computationally appropriate method, due to the innovation vectors \(\{\textbf{y}_t\}_{t=1}^T\) being white, one can relate the observation covariance matrix \(\textbf{C}_\textbf{x}=\mathbb{E}\{\textbf{x}\textbf{x}^T\}\in\mathbb{R}^{N\times N}\) with eigenvectors (\(\textbf{V}_{\diamond}\)) and eigenvalues (\(\boldsymbol{\Lambda}_{\diamond}\)) of \(\mathcal{G}_N\) via (\ref{Diffusion}) as \cite{segarra2017network}:

\begin{equation}
\label{EVDofCx}
\begin{split}
    \textbf{C}_\textbf{x}=\mathbb{E}\{\textbf{x}\textbf{x}^T\}=\textbf{V}_{\diamond}\left(\sum_{l=0}^{L-1}{h_l\boldsymbol{\Lambda}_{\diamond}^l}\right)^2\textbf{V}_{\diamond}^T
\end{split}
\end{equation}

On the other hand, from (\ref{EVDprod}), the EVD of a product graph \(\mathcal{G}_N\) can be expressed based on the EVD of its graph factors \(\mathcal{G}_P\) and \(\mathcal{G}_Q\) as:

\begin{equation}
\label{EVDofProd}
    \textbf{W}_{\diamond}=\overbrace{(\textbf{V}_P\otimes\textbf{V}_Q)}^{\textbf{V}_{\diamond}}\overbrace{(\boldsymbol{\Lambda}_P\diamond\boldsymbol{\Lambda}_Q)}^{\boldsymbol{\Lambda}_{\diamond}}\overbrace{(\textbf{V}_P\otimes\textbf{V}_Q)^T}^{\textbf{V}_{\diamond}^T}
\end{equation}

\noindent where \(\diamond\) can be \textit{any kind} of graph product, i.e., Cartesian, Kronecker, or Strong. Therefore, considering \(\textbf{x}_t=vec(\textbf{X}_t)\), where \(\textbf{X}_t=mat(\textbf{x}_t,[Q,P])\in\mathbb{R}^{Q\times P}\) denote the multi-domain expression of \(\textbf{x}_t\) for \(t=1,...,T\), eq. (\ref{Diffusion}) can be rewritten via (\ref{EVDofProd}) as:

\begin{equation}
\label{Diffusion_prod}
    vec(\textbf{X}_t)=\textbf{x}_t=(\textbf{V}_P\otimes\textbf{V}_Q)\overbrace{\left[\sum_{l=0}^{L-1}{h_l\boldsymbol{\Lambda}_{\diamond}^l}\right](\textbf{V}_P\otimes\textbf{V}_Q)^T\textbf{y}_t}^{\textbf{z}_t=vec(\textbf{Z}_t)}
\end{equation}

\noindent where \(\textbf{z}_t=vec(\textbf{Z}_t)\in\mathbb{R}^{N\times1}\) and \(\textbf{Z}_t=mat(\textbf{z}_t,[Q,P])\in\mathbb{R}^{Q\times P}\) are some intermediate variables. Afterwards, using the relation \(vec(\textbf{A}\textbf{B}\textbf{C})=(\textbf{C}^T\otimes \textbf{A})vec(\textbf{B})\) for sample matrices \(\textbf{A}\), \(\textbf{B}\) and \(\textbf{C}\) \cite{petersen2008matrix}, eq. (\ref{Diffusion_prod}) turns to:

\begin{equation}
\label{Diffusion_prod2}
    \textbf{X}_t=\textbf{V}_Q\textbf{Z}_t\textbf{V}^T_P
\end{equation}

Considering (\ref{Diffusion_prod2}) for \(T\) product graph signals \(\{\textbf{X}_t\in\mathbb{R}^{Q\times P}\}_{t=1}^T\), the mathematical expectation \(\textbf{C}_\textbf{X}=\mathbb{E}\{\textbf{X}\textbf{X}^T\}\in\mathbb{R}^{Q\times Q}\) takes the form of

\begin{equation}
\label{Cx_Cz}
    \textbf{C}_\textbf{X}=\mathbb{E}\{\textbf{X}\textbf{X}^T\}=\textbf{V}_Q\overbrace{\mathbb{E}\{\textbf{Z}\textbf{Z}^T\}}^{\textbf{C}_\textbf{Z}}\textbf{V}^T_Q
\end{equation}

\noindent where the covariance matrices \(\textbf{C}_\textbf{X}\)  and \(\textbf{C}_\textbf{Z}\) are estimated via the sample mean over the graph matrices \(\{\textbf{X}_t\}_{t=1}^T\) and \(\{\textbf{Z}_t\}_{t=1}^T\), respectively. The following theorem shows that \(\textbf{C}_\textbf{Z}\) in (\ref{Cx_Cz}) is a diagonal matrix with non-negative diagonal elements, and, therefore, due to \(\textbf{C}_\textbf{X}\) being positive semi definite, \(\textbf{V}_Q\) can be recovered by performing EVD on \(\textbf{C}_\textbf{X}=\mathbb{E}\{\textbf{X}\textbf{X}^T\}\). Similarly, \(\textbf{V}_P\) can be obtained as the eigenvectors of \(\textbf{C}'_\textbf{X}=\mathbb{E}\{\textbf{X}^T\textbf{X}\}\in\mathbb{R}^{P\times P}\).

\textbf{Theorem 2:} \textit{The covariance matrix} \(\textbf{C}_\textbf{Z}\) \textit{in} (\ref{Cx_Cz}) \textit{is diagonal with non-negative diagonal elements.}

\textit{Proof:} From (\ref{Diffusion_prod}), it can be seen that

\begin{equation}
\begin{split}
& \mathbb{E}\{\textbf{z}\textbf{z}^T\}\\
&=\left[\sum_{l=0}^{L-1}{h_l\boldsymbol{\Lambda}^l}\right](\textbf{V}_P\otimes\textbf{V}_Q)^T\overbrace{\mathbb{E}\{\textbf{y}\textbf{y}^T\}}^{\textbf{I}}(\textbf{V}_P\otimes\textbf{V}_Q)\left[\sum_{l=0}^{L-1}{h_l\boldsymbol{\Lambda}^l}\right]^T\\
&=\left[\sum_{l=0}^{L-1}{h_l\boldsymbol{\Lambda}^l}\right]\overbrace{(\textbf{V}_P\otimes\textbf{V}_Q)^T(\textbf{V}_P\otimes\textbf{V}_Q)}^{\textbf{I}}\left[\sum_{l=0}^{L-1}{h_l\boldsymbol{\Lambda}^l}\right]^T\\
&=\left[\sum_{l=0}^{L-1}{h_l\boldsymbol{\Lambda}^l}\right]^2\\
\end{split}
\end{equation}

Note that \((\textbf{V}_P\otimes\textbf{V}_Q)^T(\textbf{V}_P\otimes\textbf{V}_Q)=\textbf{I}_N\), because the orthogonality of \(\textbf{V}_P\otimes\textbf{V}_Q\) holds under the orthogonality of \(\textbf{V}_P\) and \(\textbf{V}_Q\) \cite{laub2005matrix}. Therefore, due to the diagonality of \(\mathbb{E}\{\textbf{z}\textbf{z}^T\}\), the elements of \(\textbf{z}\) are uncorrelated. On the other hand, the elements of \textbf{Z} in (\ref{Diffusion_prod}) can be described based on the elements of \(\textbf{z}\) as

\begin{equation}
\textbf{Z}=
\begin{bmatrix}
\textbf{z}_1 & \textbf{z}_{Q+1} & \hdots & \textbf{z}_{(P-1)Q+1}\\
\textbf{z}_2 & \textbf{z}_{Q+2} & \hdots & \textbf{z}_{(P-1)Q+2}\\
\vdots & \vdots & \hdots & \vdots\\
\textbf{z}_Q & \textbf{z}_{2Q} & \hdots & \textbf{z}_{PQ}
\end{bmatrix}
\end{equation}

Afterwards, \(\textbf{C}_\textbf{Z}=\mathbb{E}\{\textbf{Z}\textbf{Z}^T\}\) in (\ref{Cx_Cz}) is described as

\begin{equation}
\label{C_z}
\begin{split}
&\textbf{C}_\textbf{Z}=\\
&\begin{bmatrix}
\sum_{i=1}^{P}{\mathbb{E}\{\textbf{z}^2_{(i-1)Q+1}\}} & \hdots & \sum_{i=1}^{P}{\mathbb{E}\{\textbf{z}_{(i-1)Q+1}\textbf{z}_{iQ}\}}\\
\vdots & \ddots & \vdots\\
\sum_{i=1}^{P}{\mathbb{E}\{\textbf{z}_{iQ}\textbf{z}_{(i-1)Q+1}\}} & \hdots & \sum_{i=1}^{P}{\mathbb{E}\{\textbf{z}^2_{iQ}\}}
\end{bmatrix}
\end{split}
\end{equation}

From (\ref{C_z}), it can be seen that the diagonal elements of \(\textbf{C}_\textbf{Z}\) are the summation of some statistical variances and, therefore, are non-negative. Besides, due to the uncorrelatedness of the elements of \(\textbf{z}\), the non-diagonal elements of \(\textbf{C}_\textbf{Z}\) in (\ref{C_z}) are zero, and, therefore, \(\textbf{C}_\textbf{Z}\) is a diagonal matrix with non-negative diagonal element. $ \blacksquare $

Finally, by obtaining the factor eigenmatrices \(\textbf{V}_P\) and \(\textbf{V}_Q\), the factor graphs \(\mathcal{G}_P\) and \(\mathcal{G}_Q\) can be recovered from (\ref{obj_W}) via Algorithm \ref{alg:cap}. The mentioned proposed approach, named ProdSpecTemp, is summarized in Algorithm \ref{alg2:cap}, which we recall that can recover any product graph, without the need to known the involved graph product type.

\begin{algorithm}[!t]
\caption{: ProdSpecTemp}\label{alg2:cap}
\begin{algorithmic}[1]
\INPUT $\mathcal{X}=\{\textbf{x}_t\in\mathbb{R}^{N\times N}\}_{t=1}^{T},\: P,\: Q,\:$ where $N=PQ$
\OUTPUT Factor adjacencies $\textbf{W}_P\in\mathbb{R}^{P\times P},\:\:\:\textbf{W}_Q\in\mathbb{R}^{Q\times Q}$
\State Obtain the product descriptions $\mathcal{X}_Q=\{\textbf{X}_t=mat(\textbf{x}_t,[Q,P])\}_{t=1}^{T}$ and $\mathcal{X}_P=\{\textbf{X}'_t=\textbf{X}^T_t\}_{t=1}^{T}$
\State Estimate the covariances \(\textbf{C}_\textbf{X}=\mathbb{E}\{\textbf{X}\textbf{X}^T\}\) and \(\textbf{C}'_\textbf{X}=\mathbb{E}\{\textbf{X}^T\textbf{X}\}\) via sample mean over \(\{\textbf{X}_t\}_{t=1}^T\) and \(\{\textbf{X}^T_t\}_{t=1}^T\), respectively
\State Obtain the orthogonal factor eigenvectors \(\textbf{V}_P\) and \(\textbf{V}_Q\) via EVD on \(\textbf{C}_\textbf{X}\) and \(\textbf{C}'_\textbf{X}\)
\State Learn factor adjacencies \(\textbf{W}_p\) and \(\textbf{W}_Q\) from (\ref{obj_W}) on \(\textbf{V}_P\) and \(\textbf{V}_Q\) by performing SpecTemp-IALM (Lines 3-14 of Algorithm \ref{alg:cap})
\end{algorithmic}
\end{algorithm}

\subsection{HO-ProdSpecTemp}
To extend the domains of the higher-order graph signal \(\textbf{x}\) to more than two domains, we consider that the case in which the product graph \(\mathcal{G}_N\) has \(n\) factor graphs as \(\mathcal{G}_N=\mathcal{G}_{P_1}\diamond\mathcal{G}_{P_2}\diamond...\diamond\mathcal{G}_{P_n}\), where \(N=P_1P_2...P_n\), in the following problem as: 

\textbf{Problem 2:} \textit{Learn (any kind of Cartesian, Kronecker, or Strong) product graph \(\mathcal{G}_N=\mathcal{G}_{P_1}\diamond...\diamond\mathcal{G}_{P_n}\), where \(\{\mathcal{G}_{P_i}\}_{i=1}^n\) are its graph factors and \(N=P_1P_2...P_n\), by observing a stream of \(T\) independent stationary \(N\)-dimensional multi-domain graph signals \(\mathcal{X}=\{\textbf{x}_t\in\mathbb{R}^{N\times1}\}_{t=1}^T\) diffused on \(\mathcal{G}_N\).} 

The EVD of the product graph \(\mathcal{G}_N\) based on the EVD of its factors as

\begin{equation}
    \textbf{W}_\diamond=\overbrace{(\textbf{V}_{P_1}\otimes...\otimes\textbf{V}_{P_n})}^{\textbf{V}_\diamond}\overbrace{(\boldsymbol{\Lambda}_{P_1}\diamond...\diamond\boldsymbol{\Lambda}_{P_n})}^{\boldsymbol{\Lambda}_\diamond}\overbrace{(\textbf{V}_{P_1}\otimes...\otimes\textbf{V}_{P_n})^T}^{\textbf{V}^T_\diamond}
\end{equation}

By considering \(\underline{\textbf{X}}\in\mathbb{R}^{P_n\times P_{n-1}\times...\times P_1}\) as the higher-order diffused graph tensor representation of the observed graph signal \(\textbf{x}\in\mathbb{R}^{N\times1}\), the tensorial form of (\ref{Diffusion_prod}) can be expressed via tensorial products \cite{sidiropoulos2017tensor} as:

\begin{equation}
\label{tens}
\begin{split}
\underline{\textbf{X}}= \underline{\textbf{Z}}\times_1\textbf{V}_{P_n}\times_2\textbf{V}_{P_{n-1}}\times_3...\times_n\textbf{V}_{P_1}   
\end{split}    
\end{equation}

\noindent and the mode-\(i\) unfolding (matricization) form \cite{sidiropoulos2017tensor} of (\ref{tens}) is 

\begin{equation}
    \underline{\textbf{X}}_{(i)}=\textbf{V}_{P_{n-i+1}}\underline{\textbf{Z}}_{(i)}\left(\textbf{V}_{P_n}\otimes...\otimes\textbf{V}_{P_{i+1}}\otimes\textbf{V}_{P_{i-1}}\otimes...\otimes\textbf{V}_{P_1}\right)^T
\end{equation}

Similar to the matrix mode (\ref{Cx_Cz}), the extended higher-order form is as

\begin{equation}
\label{Cx_Cz_tensor}
    \textbf{C}_{\underline{\textbf{X}}_{(i)}}=\mathbb{E}\{\underline{\textbf{X}}_{(i)}\underline{\textbf{X}}_{(i)}^T\}=\textbf{V}_{P_i}\overbrace{\mathbb{E}\{\underline{\textbf{Z}}_{(i)}\underline{\textbf{Z}}_{(i)}^T\}}^{\textbf{C}_{\underline{\textbf{Z}}_{(i)}}}\textbf{V}^T_{P_i}
\end{equation}

Similar to the matrix mode, the following theorem helps in recovering \(\{\textbf{V}_{P_i}\}_{i=1}^n\).   

\textbf{Theorem 3:} \textit{The covariance matrix} \(\textbf{C}_{\underline{\textbf{Z}}_{(i)}}\) \textit{in} (\ref{Cx_Cz_tensor}) \textit{is diagonal with non-negative diagonal elements.}

\textit{Proof:} Similar to approach to the proof of the Theorem 1 would prove the current theorem and thus is omitted. $ \blacksquare $

Therefore, EVD on \(\textbf{C}_{\underline{\textbf{X}}_{(i)}}\) gets the \(i\)th factor eigenvector matrix \(\textbf{V}_{P_i}\). This higher-order approach (HO-ProdSpecTemp) is summarized in Algorithm \ref{alg3:cap}.

\begin{algorithm}[!t]
\caption{: HO-ProdSpecTemp}\label{alg3:cap}
\begin{algorithmic}[1]
\INPUT Tensor graph signals $\{\underline{\textbf{X}}_t\in\mathbb{R}^{P_1\times...\times P_n}\}_{t=1}^{T},\: \{P_i\}_{i=1}^n$
\OUTPUT Factor adjacencies $\{\textbf{W}_{P_i}\in\mathbb{R}^{P_i\times P_i}\}_{i=1}^n$
\For{$i=1:n$}
\State Perform \(i\)th mode unfolding on \(\{\underline{\textbf{X}}_t\}_{t=1}^{T}\) and compute 

\(\textbf{C}_{\underline{\textbf{X}}_{(i)}}=\mathbb{E}\{\underline{\textbf{X}}_{(i)}\underline{\textbf{X}}_{(i)}^T\}\)
\State Perform EVD on \(\textbf{C}_{\underline{\textbf{X}}_{(i)}}\) and obtain \(\textbf{V}_{P_{n-i+1}}\)
\State Learn factor adjacency \(\textbf{W}_{P_{n-i+1}}\) from (\ref{obj_W}) on \(\textbf{V}_{P_{n-i+1}}\) 

by performing SpecTemp-IALM (Lines 3-14 of 

Algorithm \ref{alg:cap})
\EndFor{, and return $\{\textbf{W}_{P_i}\}_{i=1}^n$}
\end{algorithmic}
\end{algorithm}

\begin{figure*}[!t]
  \centering 
  \includegraphics[width=17cm, trim={2.5cm 0cm 2.5cm 0.5cm}, clip=true]{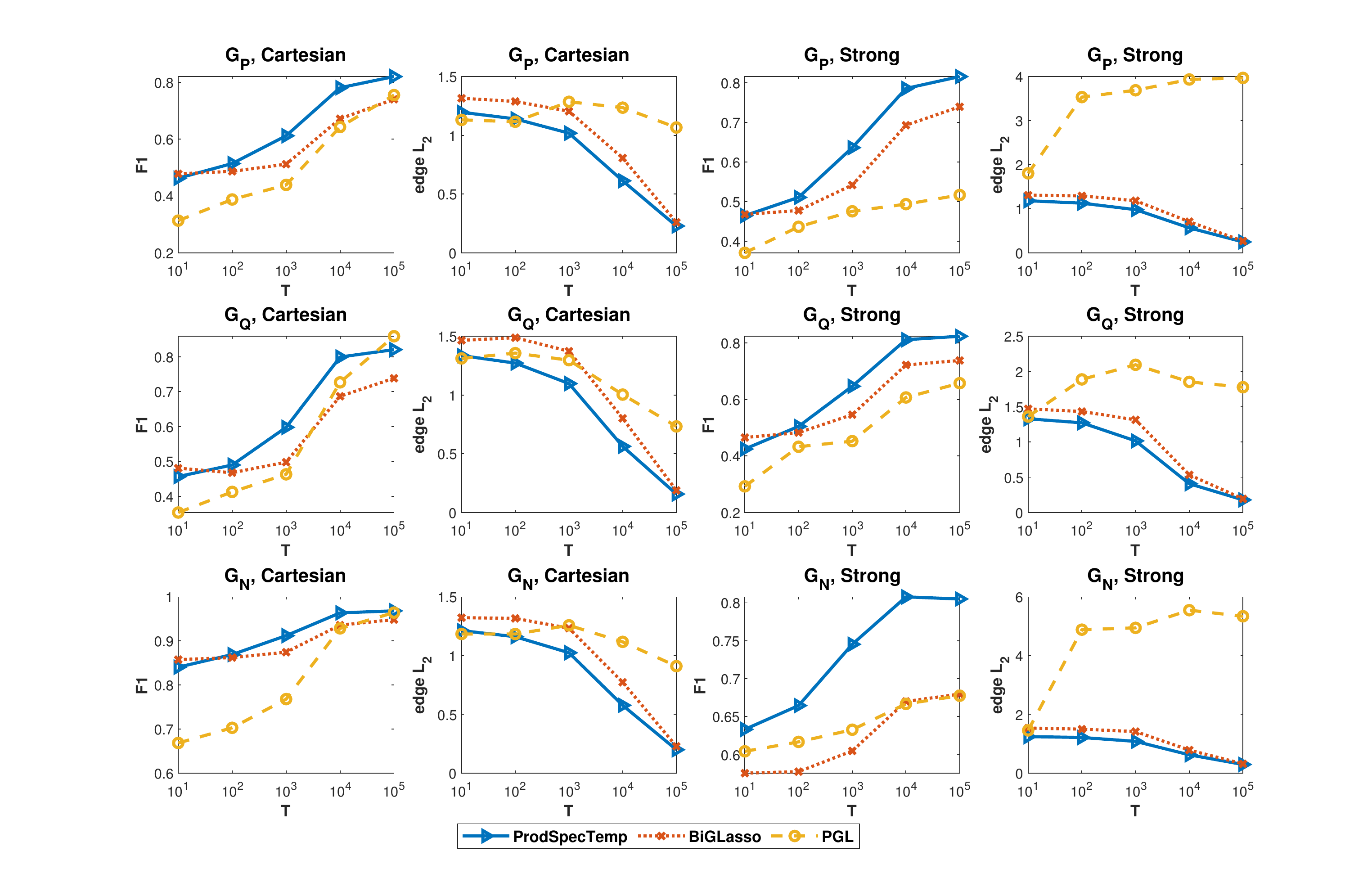} 
  \caption{The comparison of the product and factor graph learning performance of the proposed ProdSpecTemp with the BiGLasso \cite{kalaitzis2013bigraphical} and PGL \cite{kadambari2020learning,kadambari2021product}, in which the graph learning performance is evaluated via two popular metrics F1-measure \cite{dong2016learning} and edge \(\ell^{err}_2=\frac{\|\textbf{W}-\hat{\textbf{W}}\|_F^2}{\|\textbf{W}\|_F^2}\) error \cite{kalofolias2016learn}, where \(\textbf{W}\) and \(\hat{\textbf{W}}\) denote the true and learned adjacencies, respectively.}
  \label{Fig1} 
\end{figure*}

\begin{figure}[!t]
  \centering 
  \includegraphics[width=9.5cm, trim={0.5cm 0.5cm 0.5cm 0.5cm}, clip=true]{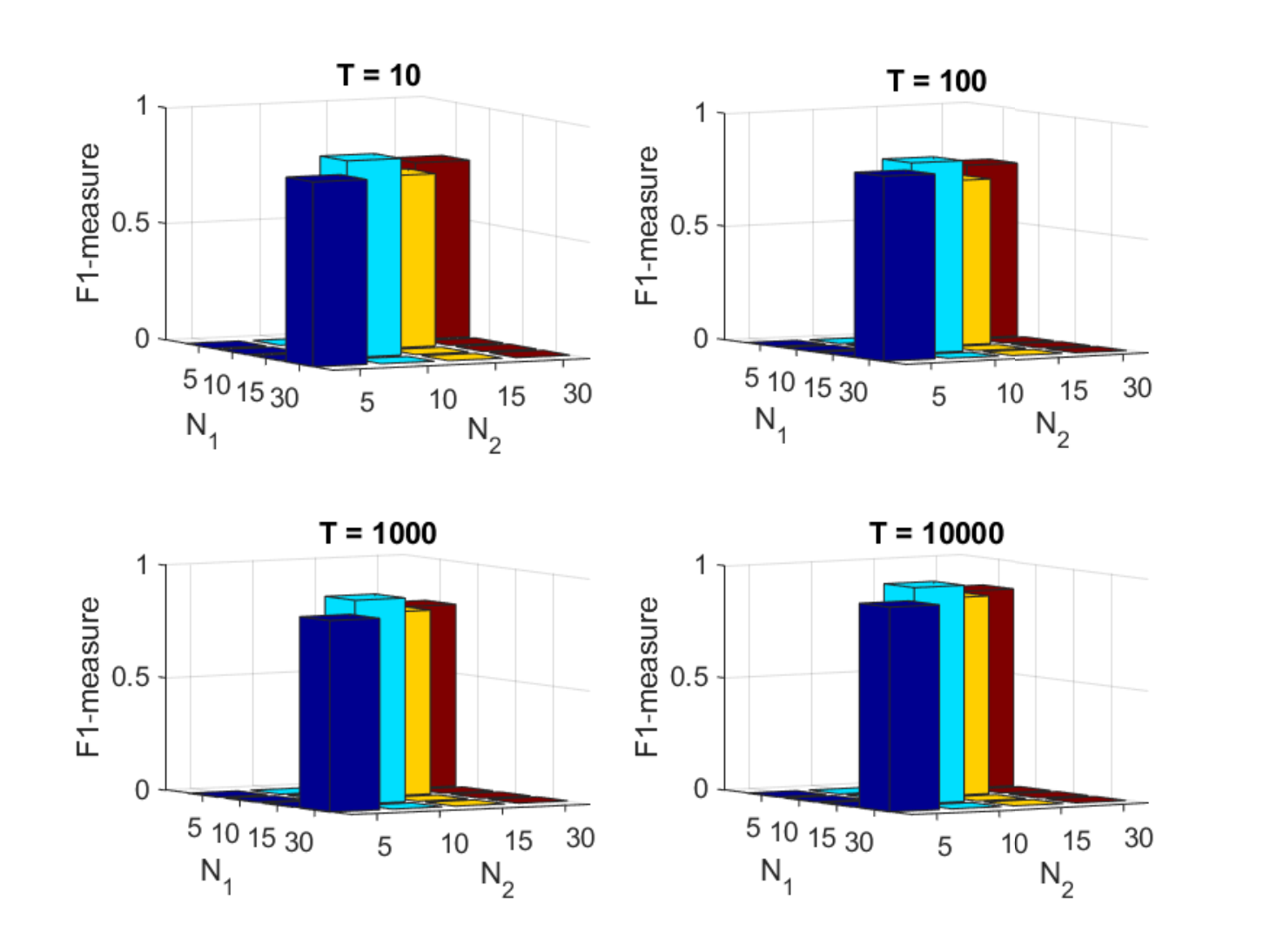} 
  \caption{The effect of unknown (not accurately estimated) \(P\) and \(Q\) investigated by the F1-measure of the resulted product graph averaged over 20 noisy independent realizations (\(\text{SNR}=-20\)db) across \(N_1\) and \(N_2\) and different number of graph signals \(T\), where \(P_{true}=15\) and \(Q_{true}=10\).}
  \label{Hyperparam} 
\end{figure}

\begin{figure}[!t]
  \centering 
  \includegraphics[width=9.5cm, trim={1cm 0cm 0cm 0cm}, clip=true]{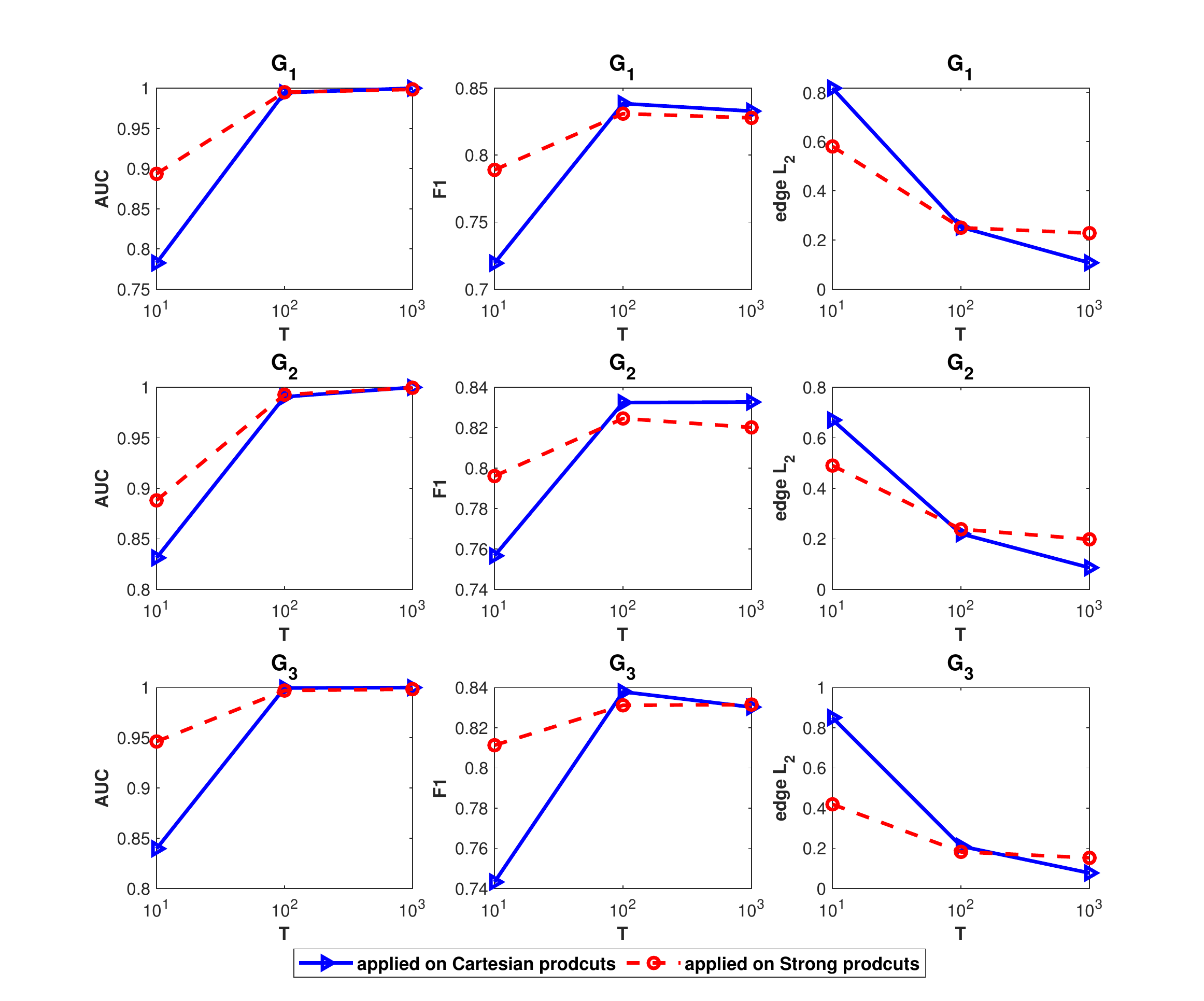} 
  \caption{The average of the graph recovery performance corresponding to the three factor graphs over twenty independent realizations.}
  \label{Fig2} 
\end{figure}

\begin{figure*}[!t]
  \centering 
  \includegraphics[width=18.5cm, trim={5cm 1.5cm 0cm 0cm}, clip=true]{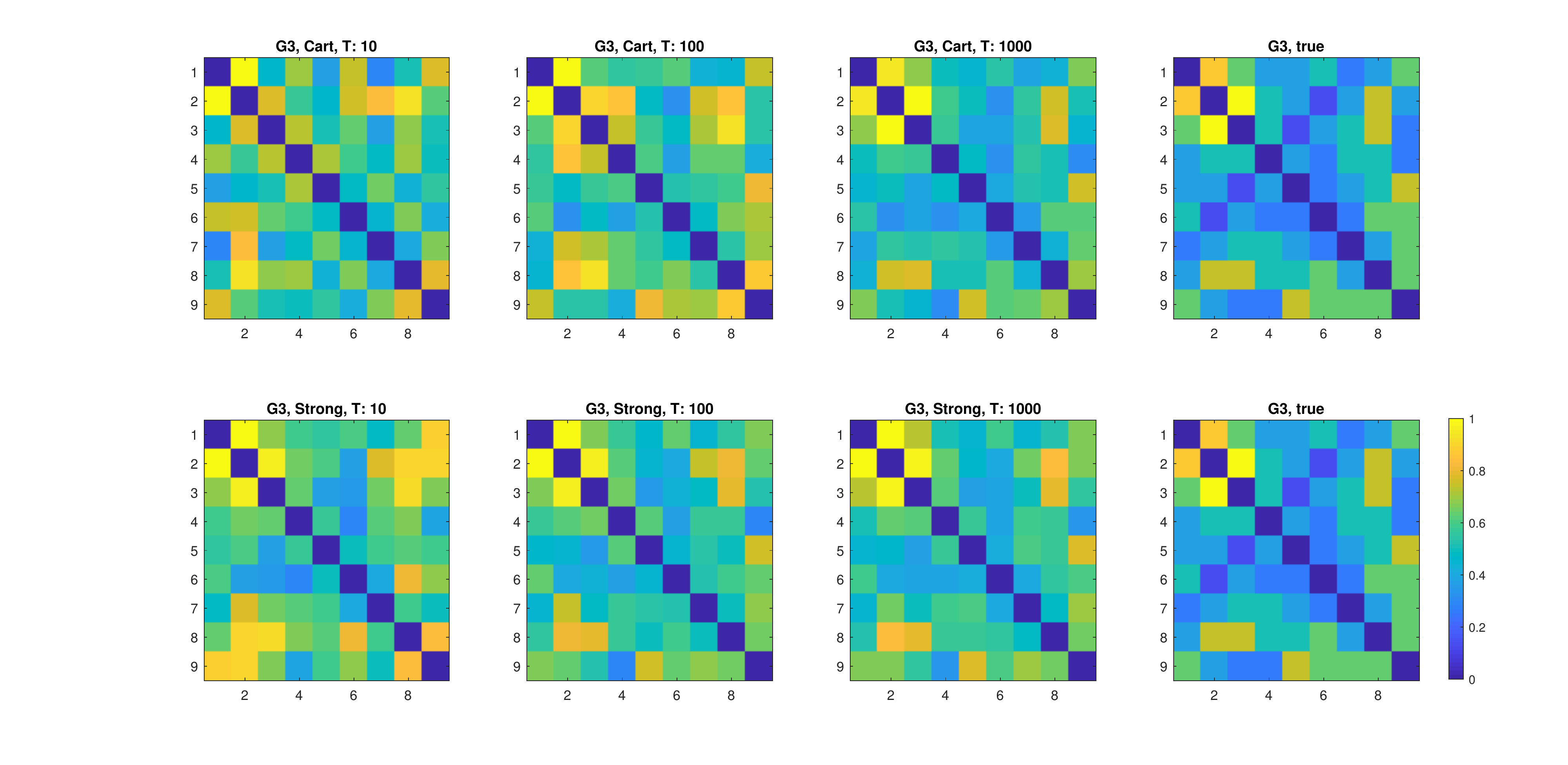} 
  \caption{The average of the true and third learned factor graphs (\(N_3=9\)) from Cartesian (top row) and Strong (bottom row) product graph signals.}
  \label{Fig3} 
\end{figure*}


\subsection{Computational Complexity Analysis}
\label{ComplexSec}
It can be seen that performing HdSpecTemp on (\ref{obj_W}) optimizes \(N^2+N=P^2Q^2+PQ\) (in tensor mode, \(N^2+N=\Pi_{i=1}^{n}{P^2_i}+\Pi_{i=1}^{n}{P_i}\)) free variables; however, ProdSpecTemp considers two (in tensor mode, \(n\ge3\)) completely separable GL problems on the graph factors \(\textbf{W}_P\) and \(\textbf{W}_Q\) (in tensor mode, \(\{\textbf{W}_{P_i}\}_{i=1}^n\)) separately, which reduces the number of optimization variables to \(P^2+Q^2+P+Q\) (in tensor mode, \(\sum_{i=1}^{n}{P^2_i}+\sum_{i=1}^{n}{P_i}\)). This number can be even more reduced to \(\frac{P(P-1)}{2}+\frac{Q(Q-1)}{2}+P+Q\) (in tensor mode, \(\sum_{i=1}^{n}{\frac{P_i(P_i-1)}{2}}+\sum_{i=1}^{n}{P_i}\)) by exploiting the proposed SpecTemp-IALM (Algorithm \ref{alg:cap}) in the mentioned separated optimizations. 

Besides, it has been shown \cite{segarra2017network} that the computational complexity of (\ref{obj_W}) is dominated by EVD of observation covariance matrix, which requires \(\mathcal{O}(N^3)=\mathcal{O}(P^3Q^3)\)  (in tensor mode, \(\mathcal{O}(N^3)=\mathcal{O}(\Pi_{i=1}^{n}{P_i^3})\)) operations. However, due to the separability of recovering factor graphs in the proposed Algorithm \ref{alg2:cap}, the proposed approach requires significantly reduced \(\mathcal{O}(P^3+Q^3)\) (in tensor mode, \(\sum_{i=1}^{n}{P_i^3})\)) operations.

\section{Experimental Results and Discussion}
\label{Sec7}
\subsection{Comparison to the Related Work}
In this subsection, the proposed ProdSpecTemp method is compared with the related methods BiGLasso \cite{kalaitzis2013bigraphical} and PGL \cite{kadambari2021product} for learning Erdös-Rényi (ER) factor graphs \(\mathcal{G}_P\) and \(\mathcal{G}_Q\) (\(P=15, \:Q=10,\:N=PQ=150\)) with the edge probability \(p_{\text{\tiny ER}}=0.3\) simulated by GSPBOX \cite{perraudin2014gspbox}. \(T\) product graph signals (\(T \in \{10,10^2,10^3,10^4,10^5\}\)) are generated from Cartesian and Strong product graph diffusion processes (with \(L=2\), \(h_0=1,\:h_1=0.5)\), and innovation vectors \(\{\textbf{y}_t\sim\mathcal{N}(\textbf{0}_N,\textbf{I}_N)\}_{t=1}^{T}\) in (\ref{Diffusion})). To make the settings more challenging and also investigate the asymptotic behaviour of the involved methods on noisy observations, we add Gaussian noise with Signal to Noise Ratio \((\text{SNR})=-20\)db to the resulted product graph signals, and the product and factor graph learning results are illustrated in Figure \ref{Fig1}, in which the graph learning performance is evaluated via two popular metrics F1-measure \cite{dong2016learning} and edge \(\ell^{err}_2=\frac{\|\textbf{W}-\hat{\textbf{W}}\|_F^2}{\|\textbf{W}\|_F^2}\) error \cite{kalofolias2016learn}, where \(\textbf{W}\) and \(\hat{\textbf{W}}\) denote the true and learned adjacencies, respectively. As can be seen in this figure, with increasing the number of graph signals (\(T\)) at hand, the performances in (almost) all cases improve. Due to the suitability of the BiGLasso \cite{kalaitzis2013bigraphical} and PGL \cite{kadambari2021product} methods for recovering factor graphs from Cartesian products (but with different specific assumptions), their performances corresponding to the Cartesian products are better than Strong ones. However, the proposed ProdSpecTemp has superior performance over the compared methods in recovering both the Cartesian and Strong product graphs. Besides, the rather robustness of the ProdSpecTemp method against a large amount of noise is verified. 
%

\subsection{Hyperparameter Analysis (\(P\) and \(Q\))}

First, we recall that the only notable hyperparameter of the proposed ProdSpecTemp method is the number of nodes of the factor graphs, i.e., \(P\) and \(Q\). To analyze the effect of unknown \(P\) and \(Q\) and also estimate them in the case of having ground truth product graphs, Figure \ref{Hyperparam} shows the F1-measure of the resulted product graph averaged over 20 noisy independent realizations (\(\text{SNR}=-20\)db) across \(N_1\) and \(N_2\) in the span of \(\in\{5,10,15,30\}\) and different number of graph signals \(T\in\{10^1,10^2,10^3,10^4\}\), where \(P=15\) and \(Q=10\). Note that, in this figure, the valid points must satisfy \(N_1\times N_2=P \times Q=150\), and, therefore, the invalid points take the zero value. As can be seen in this figure, the point \((N_1,N_2)=(P,Q)=(15,10)\) has the highest F1-measure in all values of \(T\). Besides, even in the points \((N_1,N_2)\ne(P,Q)\), the F1-measure does not drop drastically, especially in the case of having a fair number of graph signals, e.g., \(T=1000\). This observation implies the robustness of the proposed method against the unknown (or not accurately estimated) hyperparameters. 

\begin{figure*}[!t]
  \centering 
  \includegraphics[width=19.5cm, trim={5cm 1.5cm 0cm 0cm}, clip=true]{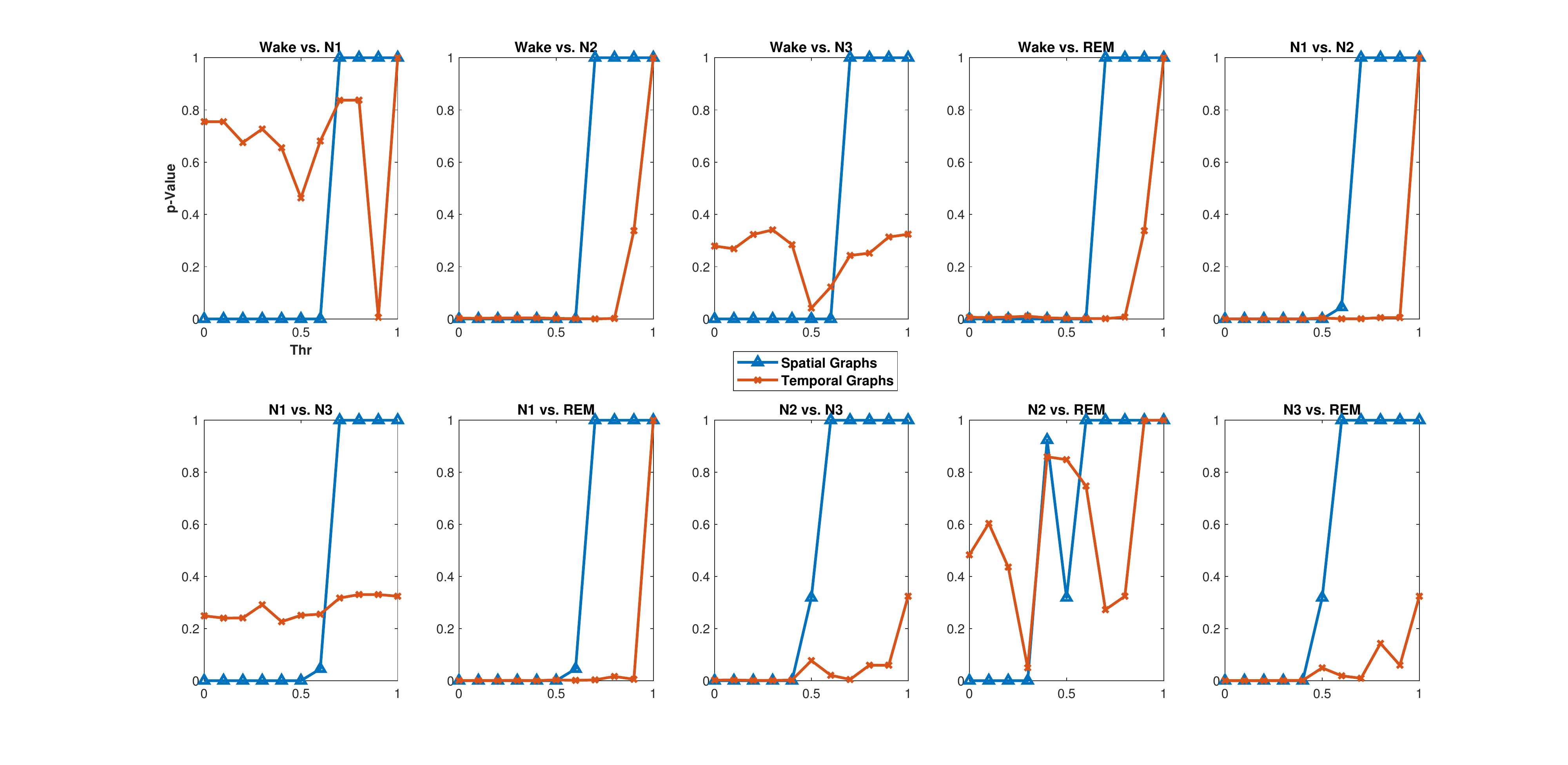} 
  \caption{The \(p\)-values obtained from \(t\)-test in the threshold span of \(\{0:0.1:1\}\) corresponding to the pairwise sleep stages.}
  \label{Fig4} 
\end{figure*}

\begin{figure}[!t]
  \centering 
  \includegraphics[width=9cm, trim={0cm 0cm 0cm 0cm}, clip=true]{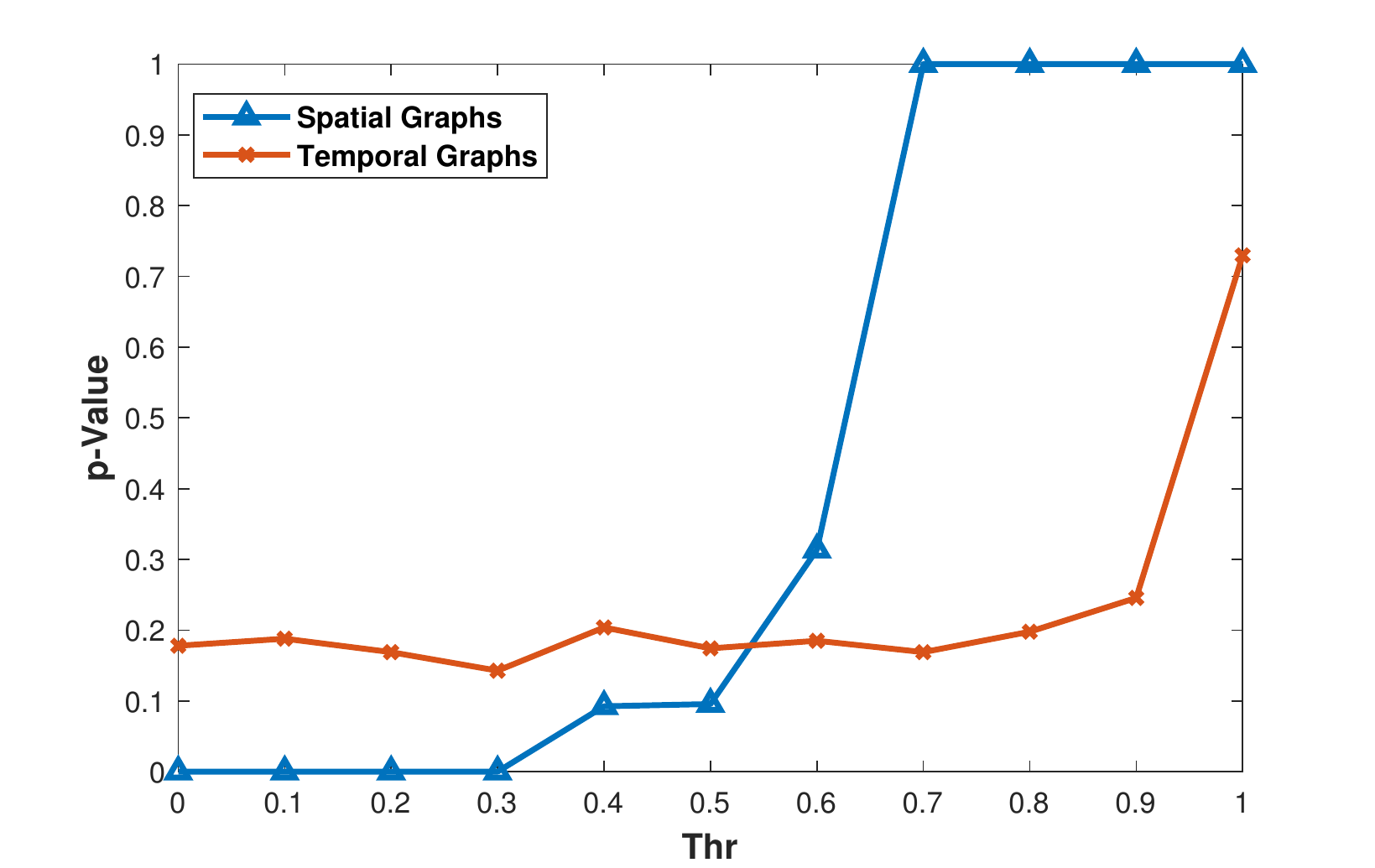}
  \caption{The average of the \(p\)-values of the threshold span corresponding to Figure \ref{Fig4}.}
  \label{Fig5} 
\end{figure}

\begin{figure*}[!t]
  \centering 
  \includegraphics[width=16.5cm, trim={5cm 1.2cm 4cm 1cm}, clip=true]{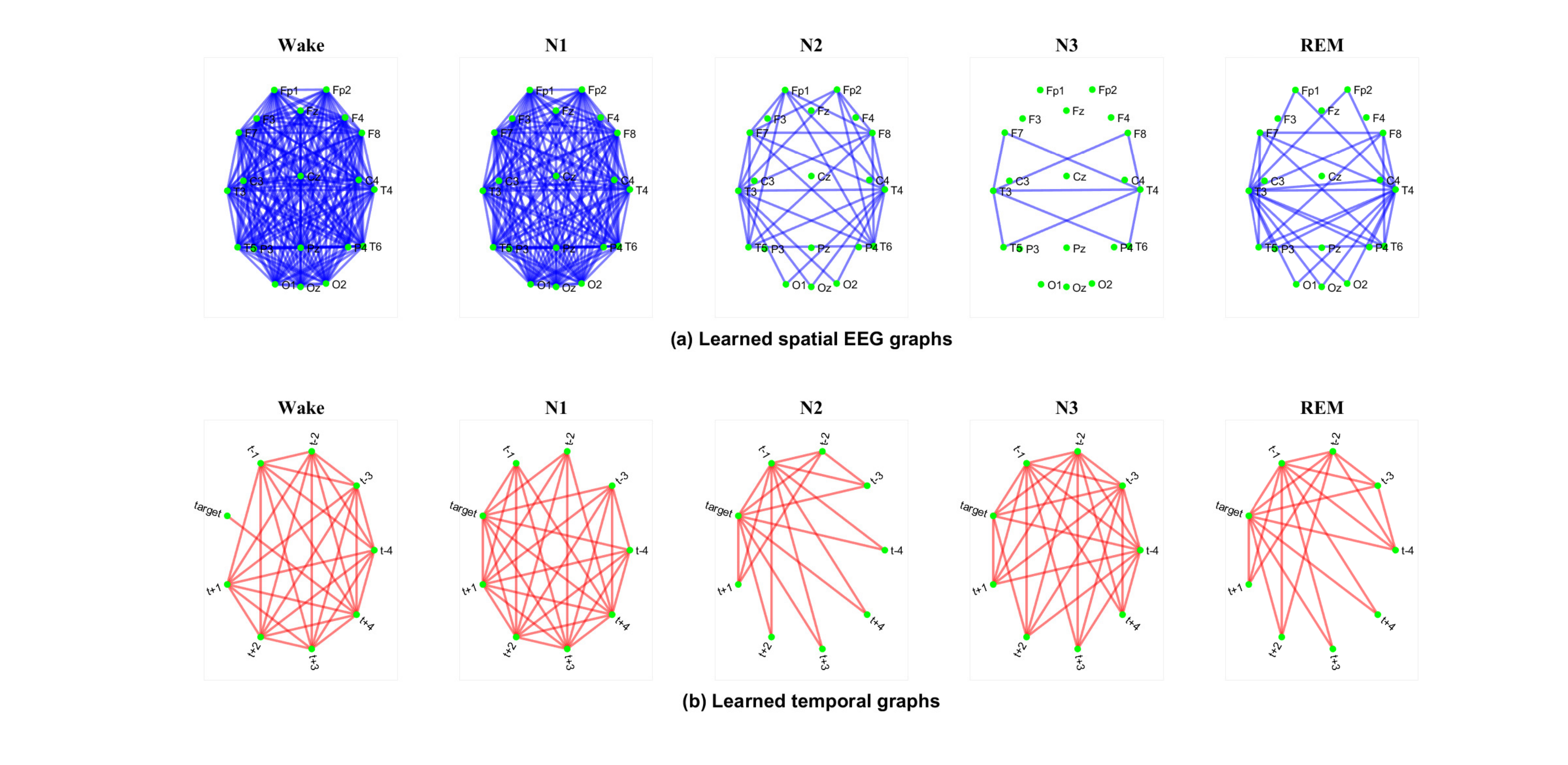} 
  \caption{a) The averaged learned spatial graphs related to the EEG channels, b) the averaged learned temporal graphs from neighbor thirty-second sleep epochs.}
  \label{Fig6} 
\end{figure*}

\begin{figure}[!t]
  \centering 
  \includegraphics[width=9cm, trim={2cm 0cm 0cm 0.3cm}, clip=true]{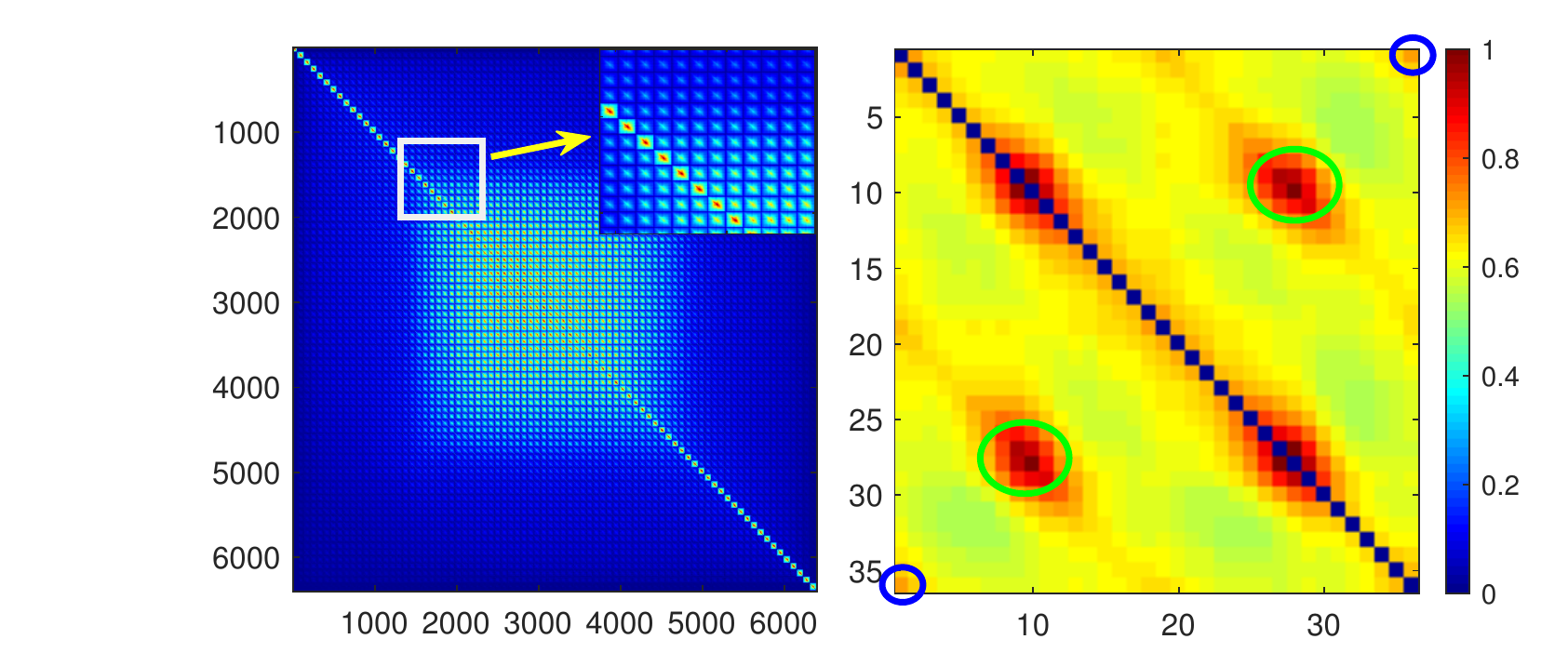} 
  \caption{The learned \(6400\times6400\) and \(36\times36\) adjacencies corresponding to the pixel (left panel) and degree (right panel) graphs, respectively.}
  \label{Fig7} 
\end{figure}

\begin{figure}[!t]
  \centering 
  \includegraphics[width=8cm, trim={0cm 1cm 0cm 0cm}, clip=true]{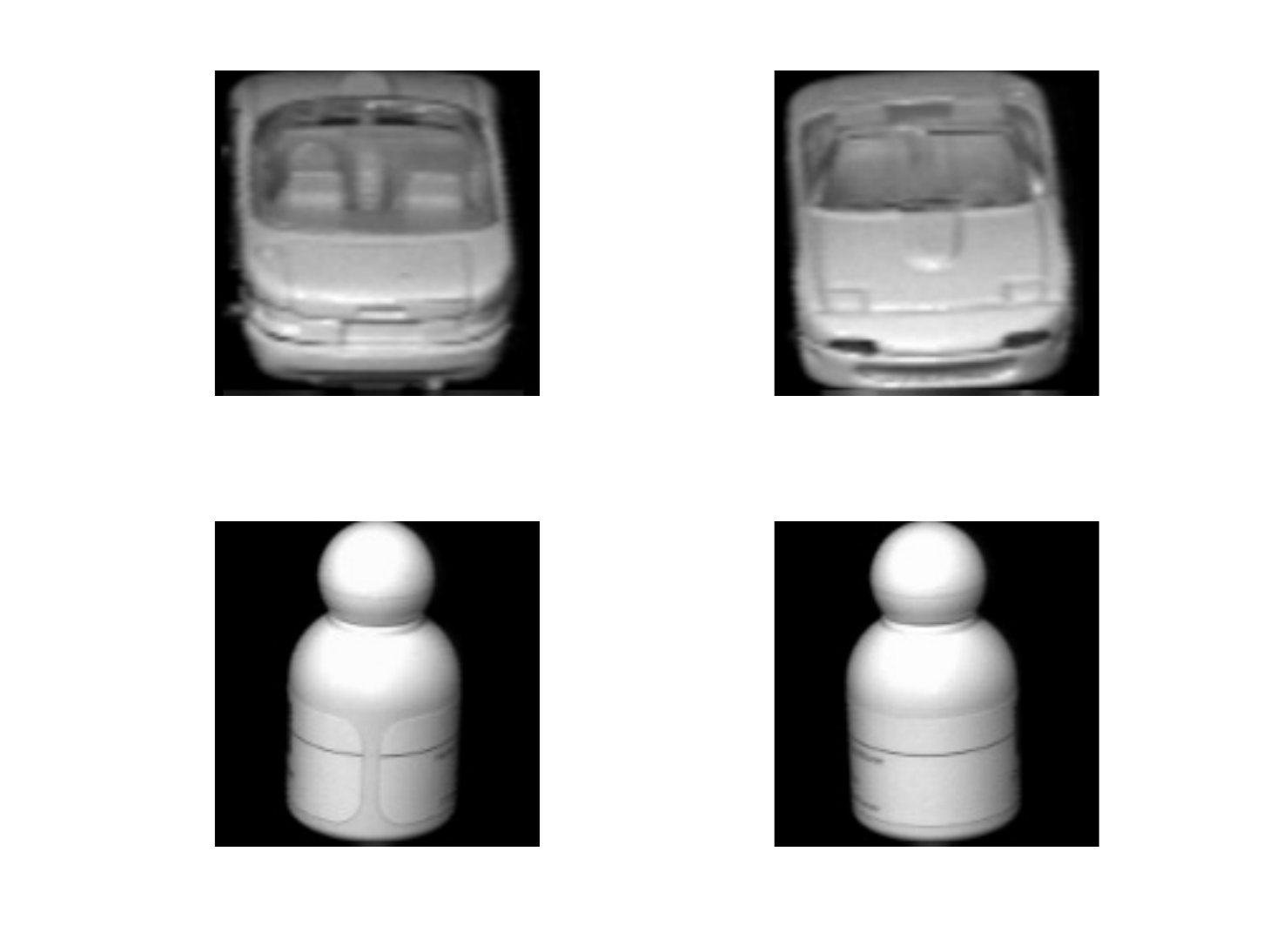} 
  \caption{Two examples of the cases in which the objects are quite in front and back view positions.}
  \label{Similar_degrees} 
\end{figure}

\begin{figure}[!t]
  \centering 
  \includegraphics[width=8.5cm, trim={3.5cm 2cm 1cm 1cm}, clip=true]{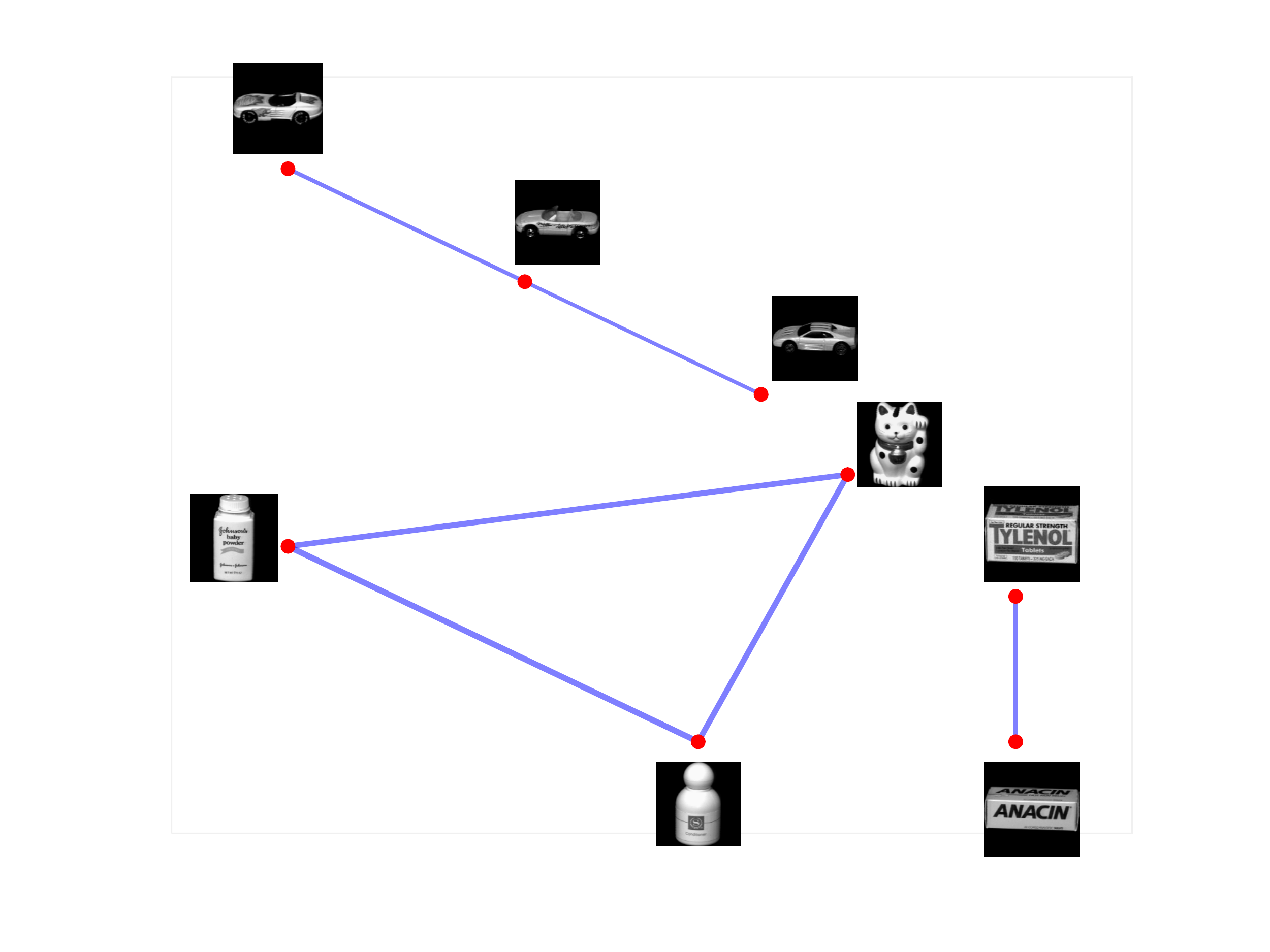} 
  \caption{The learned object graph, which shows the meaningful connectivities between similar objects and in similar poses, e.g., cars.}
  \label{Fig8} 
\end{figure}

\subsection{Learning more than Two Factor Graphs}
To illustrate the flexibility of the proposed HO-ProdSpecTemp method in recovering more than two factor graphs, we generate three factor ER graphs with \(P_1=12,\:\: P_2=10\) and \(P_3=9\) number of nodes and with the edge probability \(p_{\text{\tiny ER}}=0.3\) for making Cartesian and Strong product graphs \(\mathcal{G}_N\). Then, \(T\) product graph signals, where \(T\in\{10, 10^2, 10^3\}\), are simulated via (\ref{Diffusion}) (with \(L=2\), \(h_0=1,\:h_1=0.5)\), and innovation vectors \(\{\textbf{y}_t\sim\mathcal{N}(\textbf{0}_N,\textbf{I}_N)\}_{t=1}^{T}\). Afterward, the Gaussian additive noise with \(\text{SNR}=-10\)db is added to the generated graph signals. The average of the graph recovery performance corresponding to the three factor graphs over twenty independent realizations are illustrated in Figure \ref{Fig2} (with added Area Under Curve (AUC) metric), which shows that, although recovery of the Strong products is more challenging than the Cartesian ones, the proposed HO-ProdSpecTemp method has successfully recovered the factor graphs with improved performance in the case of the increasing number of observed graph signals \(T\). Besides, the average of the true and third learned factor graphs (\(N_3=9\)) from Cartesian and Strong product graph signals are provided in Figure \ref{Fig3}, which supports the success in graph recovery results and deductions in Figure \ref{Fig2}.

\subsection{Learning Brain Functional Connectivity of Sleep Stages}

To investigate the applicability of the proposed ProdSpecTemp on real-world Electroencephalogram (EEG) brain signals, we select the MASS-SS3 dataset \cite{o2014montreal} consisting of Polysomnogram (PSG) data from 62 healthy subjects during sleep, i.e., 20 EEG, 3 Electromyogram (EMG), 2 Electrooculogram (EOG), and 1 Electrocardiogram (ECG) channels with a sampling frequency of 256 Hz. The sleep scoring of the thirty-second sleep epochs into five sleep stages (Wake, Rapid Eye Movement (REM), N1, N2, and N3) is performed by sleep experts using the American Academy of Sleep Medicine (AASM) standard \cite{iber2007aasm}. With no specific preprocessing on the PSG signals, Differential Entropy (DE) features in 9 crossed frequency bands 0.5-4 Hz, 2-6 Hz, 4-8 Hz, 6-11 Hz, 8-14 Hz, 11-22 Hz, 14-31 Hz, 22-40 Hz, and 31-50 Hz were extracted from each of PSG channels. As well as the spatial EEG electrodes' connectivities having 26 spatial factor graph nodes, the temporal connections of neighbor thirty-second sleep epochs have also been reported in the relevant literature for efficient sleep staging \cite{supratak2017deepsleepnet,jiang2019robust,chambon2018deep}, and, therefore, we consider four epochs before and after the target epoch and, therefore, temporal factor graphs have nine nodes. 

It has been shown that analysis of the spatial graphs can represent valuable information about brain functional connectivity during sleep \cite{alper2013weighted, huang2018graph}, where Nguyen et al. \cite{nguyen2018exploring} showed that the functional connectivity of the brain varies across different sleep stages. In this way, we plot and analyze the averaged sparsified learned spatial/temporal graphs based on a specific threshold. 

To determine the thresholds that lead to the most significant differences, we propose a statistical procedure in which the thresholds are specified based on the statistical significance difference between edge values of different sleep stage graphs. We calculate the \(p\)-values obtained from a statistical test, e.g., \(t\)-test, in the threshold span of \(\{0:0.1:1\}\) corresponding to the pairwise sleep stages in Figure \ref{Fig4}. Besides, the average of the \(p\)-values of the threshold span corresponding to the graph connectivities is illustrated in Figure \ref{Fig5}. From these results, it can be seen that \(Thr=0.4\) and \(Thr=0.3\) have approximately led to the lowest (non-zero) \(p\)-value and consequently higher statistical significance between different sleep stages for spatial and temporal averaged graphs, respectively. Therefore, these obtained thresholds are set to illustrate the binary averaged graphs.  

Figure \ref{Fig6} (a), i.e., the averaged learned spatial graphs related to the EEG channels, shows that during non-REM sleep stages (i.e., N1-3), the brain connectivities decrease compared to the Wake state, which is consistent with the findings of previous studies admitting reduced hypothalamic functional connectivity, which may be synchronized for establishing and maintaining sleep \cite{kaufmann2006brain, tarun2021nrem}. Moreover, this figure shows an increase in connectivity and activity of the occipital region corresponding to the REM vs. non-REM, which is consistent with the results of \cite{kjaer2002regional, kaufmann2006brain} that illustrate the occipital metabolism and, therefore, its connectivity with the other brain regions increases during REM vs. non-REM. Besides, the N1 stage has the most connection-based activity vs. N2 and N3, and also N3 has the lowest, which are quite consistent and supported by the neuroscientific research literature \cite{jia2021multi,larson2011modulation} implying that the N1 stage is a complicated stage and the brain is still active unlike the N3 stage which is a typical deep stage and the brain is in its low active mode.  

Figure \ref{Fig6} (b) shows the averaged learned temporal graphs. It can be seen that the target epoch is (mostly) connected with its before/after epochs; however, the key point of this figure is that even distant neighboring epochs (e.g., \(t-4\) and \(t+3\)) do not necessarily follow a tree-like graph structure, which means that different sleep epochs could have connections, and it is not limited to only before/after neighbor epochs. 

\subsection{COIL-20 Dataset}
As another real-world application of the proposed methods, in this subsection, we consider the COIL-20 dataset \cite{nenecolumbia} consisting of grayscale images (with a size of \(128\times128\) pixels) of 20 objects captured on a 5-degree interval of a turntable in front of a fixed camera. For this experiment, we select eight objects and 10-degree intervals, resulting in 36 degree images per object. Besides, we downsample the images to the size of \(80\times80\) via the bicubic interpolation approach. The mentioned structure of the data at hand allows us to consider the whole dataset as a four-way tensor \(\underline{\textbf{X}}\in\mathbb{R}^{8\times36\times80\times80}\). Therefore, we seek to learn meaningful and interpretable object, degree, and pixel connectivities by performing the proposed HO-ProdSpecTemp on \(\underline{\textbf{X}}\), where \(T=1\) in this experiment. 

The learned \(6400\times6400\) and \(36\times36\) adjacencies corresponding to the pixel and degree graphs are illustrated in the left and right panels of Figure \ref{Fig7}, respectively. Note that the shown pixel graph is the Strong product of the learned 80-node row and column graphs, as mentioned to be appropriate and comprehensive to model the pixel connectivities in image processing literature \cite{sandryhaila2014big}. As can be seen in this figure, the resulted pixel graph is approximately divided into \(80\times80\) strong connection intervals (roughly), implying the actual size of the images. On the other hand, in the resulted degree graph, notable tree structure connections along the main diagonal rely on the connections between before/after degree images. The blue circles show the connection between zero and 360-degree turns, which is quite expected. Another interesting point is the connections corresponding to the green circle area, which is associated with the cases in which the objects are quite in front and back view positions. Two examples of this scenario are shown in Figure \ref{Similar_degrees}. These findings are consistent and supported by previous pioneer work on this kind of data, e.g., \cite{kalaitzis2013bigraphical}.    

In Figure \ref{Fig8}, the learned object graph is illustrated, showing the meaningful connectivities between similar objects and in similar poses, e.g., cars. This shows the opportunity to modify the proposed methods to learn rank-constrained structures (Laplacians) for use in clustering applications (Multi-View Object Clustering) in future work.

\section{Conclusion}
\label{Sec8}

In this paper, we proposed GL approaches inferring product graphs from spectral templates of high-dimensional graph signals with possibly more than two factor graphs with significantly reduced computational complexity than the basic approach. Our approach is not limited to the specific type of graph product, in contrast to the current approaches for inference from smooth graph signals and only specific Cartesian graph products. In addition to outperforming the currently limited approaches in the synthetic diffused stationary graph signals, our approach also infers (possibly more than two) meaningful and interpretable factor graphs from sleep brain signals and multi-view object images, which are supported by expert-related pioneer previous work.

\bibliographystyle{unsrt}
\bibliography{Paper}

\clearpage
\section{Appendix}
\subsection{Simplifications of optimization (\ref{obj_w})}
\label{Simp}
The constraint \(\textbf{W}=\textbf{V} \boldsymbol{\Lambda} \textbf{V}^\text{T}\), where \(\textbf{V}=(\textbf{v}_1,...,\textbf{v}_N)^T\), can be rewritten using the vectorized operator \(vec\) as:

\begin{equation}
			\begin{aligned}
				\begin{split}
					vec(\textbf{W}) & =vec(\textbf{V} \boldsymbol{\Lambda} \textbf{V}^\text{T})=vec(\lambda_1\textbf{v}_1\textbf{v}_1^\text{T}+...+\lambda_N\textbf{v}_N\textbf{v}_N^\text{T})\\
					& = \overbrace{[vec(\textbf{v}_1\textbf{v}_1^\text{T})|...|vec(\textbf{v}_N\textbf{v}_N^\text{T})]}^{\tilde{\textbf{V}}=\textbf{V}\odot\textbf{V}}\overbrace{[\lambda_1,...,\lambda_N]^\text{T}}^{\boldsymbol{\lambda}_p}
				\end{split}
			\end{aligned}
\end{equation}

Also, vectorized form of the \(\textbf{W}\) can be written based on its upper triangular form \(\textbf{w}\in\mathbb{R}^{\frac{N(N-1)}{2} \times 1}\) as:

\begin{equation}
			\begin{aligned}
				\begin{split}
					& vec(\textbf{W})=\textbf{M}_{d}vech(\textbf{W})=\textbf{M}_{d}\textbf{M}_hvechn(\textbf{W})=\textbf{M}_{d}\textbf{M}_h\textbf{w}\\
				\end{split}
			\end{aligned}
\end{equation}

\noindent where, \(\textbf{M}_{d}\) and \(\textbf{M}_h\) are the duplication matrix \cite{abadir2005matrix} and a matrix that \(\textbf{M}_hvechn(\textbf{Z})=vech(\textbf{Z})\), for a sample zero diagonal matrix \(\textbf{Z}\), respectively. Therefore, using combination of the two previous equations, the constraint \(\textbf{W}=\textbf{V} \boldsymbol{\Lambda} \textbf{V}^\text{T}\) can be turned into:

\begin{equation}
			\textbf{w}=\boldsymbol{\Phi}\boldsymbol{\lambda}
\end{equation}

\noindent where

\begin{equation}
			\boldsymbol{\Phi} = (\textbf{M}_{d}\textbf{M}_h)^{\dagger}\Tilde{\textbf{V}}
\end{equation}

\subsection{Derivations of iteration updates of Lagrangian function (\ref{lagr})}
\label{ItUpdt}

From the Lagrangian function (\ref{lagr}), the update steps of the \((k+1)\)th iteration can be expressed as:

\begin{equation}
\begin{split}
	& \textbf{w}^{(k+1)}=\\
	&\argmin_{\textbf{w}}{\left[ \|\textbf{w}\|_1+\frac{\rho^{(k)}}{2}\left(\|\textbf{w}\|^2_2-2\langle \textbf{w}, \boldsymbol{\Phi}\boldsymbol{\lambda}^{(k)}\rangle\right)+ \frac{\rho^{(k)}}{2}\left(\|\textbf{w}\|_2^2-2\langle \textbf{w},\textbf{s}^{(k)}\rangle\right)-\langle \textbf{w},\boldsymbol{\gamma}^{(k)}_1\rangle-\langle \textbf{w},\boldsymbol{\gamma}^{(k)}_2\rangle\right]} \\
	&=\argmin_{\textbf{w}}{\left[\|\textbf{w}\|_1 +\rho^{(k)}\left(\|\textbf{w}\|_2^2-2\langle \textbf{w},\frac{\rho^{(k)}\boldsymbol{\Phi}\boldsymbol{\lambda}^{(k)}+\rho^{(k)}\textbf{s}^{(k)}+\boldsymbol{\gamma}^{(k)}_1+\boldsymbol{\gamma}^{(k)}_2}{2\rho^{(k)}}\rangle\right)\right]} \\
	&=\argmin_{\textbf{w}}{\|\textbf{w}\|_1+\rho^{(k)}\left(\left\|\textbf{w}-\frac{\rho^{(k)}\boldsymbol{\Phi}\boldsymbol{\lambda}^{(k)}+\rho^{(k)}\textbf{s}^{(k)}+\boldsymbol{\gamma}^{(k)}_1+\boldsymbol{\gamma}^{(k)}_2}{2\rho^{(k)}}\right\|_2^2\right)} \\
	&=\argmin_{\textbf{w}}{\frac{\|\textbf{w}\|_1}{2\rho^{(k)}}+\frac{1}{2}\left(\left\|\textbf{w}-\frac{\rho^{(k)}\boldsymbol{\Phi}\boldsymbol{\lambda}^{(k)}+\rho^{(k)}\textbf{s}^{(k)}+\boldsymbol{\gamma}^{(k)}_1+\boldsymbol{\gamma}^{(k)}_2}{2\rho^{(k)}}\right\|_2^2\right)} \\
	&=prox_{\frac{\|\textbf{w}\|_1}{2\rho^{(k)}}}{\frac{1}{2}\left(\left\|\textbf{w}-\frac{\rho^{(k)}\boldsymbol{\Phi}\boldsymbol{\lambda}^{(k)}+\rho^{(k)}\textbf{s}^{(k)}+\boldsymbol{\gamma}^{(k)}_1+\boldsymbol{\gamma}^{(k)}_2}{2\rho^{(k)}}\right\|_2^2\right)} \\
\end{split}
\end{equation}

\begin{equation}
\begin{split}
&\boldsymbol{\lambda}^{(k+1)}=\argmin_{\boldsymbol{\lambda}}{\frac{\rho^{(k)}}{2}\|\boldsymbol{\Phi}\boldsymbol{\lambda}\|_2^2-\langle \boldsymbol{\Phi}\boldsymbol{\lambda},\rho^{(k)}\textbf{w}^{(k+1)}\rangle+\langle \boldsymbol{\Phi}\boldsymbol{\lambda},\boldsymbol{\gamma}^{(k)}_1\rangle}\\
&=\argmin_{\boldsymbol{\lambda}}{\frac{\rho^{(k)}}{2}\left(\|\boldsymbol{\Phi}\boldsymbol{\lambda}\|_2^2-2\langle \boldsymbol{\Phi}\boldsymbol{\lambda},\frac{\rho^{(k)}\textbf{w}^{(k+1)}-\boldsymbol{\gamma}^{(k)}_1}{\rho^{(k)}}\rangle\right)}\\
&=\argmin_{\boldsymbol{\lambda}}{\frac{\rho^{(k)}}{2}\left\|\boldsymbol{\Phi}\boldsymbol{\lambda}-\frac{\rho^{(k)}\textbf{w}^{(k+1)}-\boldsymbol{\gamma}^{(k)}_1}{\rho^{(k)}}\right\|_2^2}\\
&=\boldsymbol{\Phi}^{\dagger}\left(\frac{\rho^{(k)}\textbf{w}^{(k+1)}-\boldsymbol{\gamma}^{(k)}_1}{\rho^{(k)}}\right)
\end{split}    
\end{equation}

\begin{equation}
\begin{split}
& \textbf{s}^{(k+1)}=\argmin_{\textbf{s}\in\mathcal{W}_r}{\frac{\rho^{(k)}}{2}\|\textbf{s}\|_2^2-\rho^{(k)}\langle \textbf{s},\textbf{w}^{(k+1)}\rangle+\langle \textbf{s},\boldsymbol{\gamma}^{(k)}_2\rangle}\\
&= \argmin_{\textbf{s}\in\mathcal{W}_r}{\frac{\rho^{(k)}}{2}\|\textbf{s}\|_2^2-\left\langle \textbf{s},\rho^{(k)}\textbf{w}^{(k+1)}-\boldsymbol{\gamma}^{(k+1)}_2\right\rangle}\\
&= \argmin_{\textbf{s}\in\mathcal{W}_r}{\|\textbf{s}\|_2^2-2\langle \textbf{s},\frac{\rho^{(k)}\textbf{w}^{(k+1)}-\boldsymbol{\gamma}^{(k)}_2}{\rho^{(k)}}\rangle}\\
&=\argmin_{\textbf{s}\in\mathcal{W}_r}{\frac{\rho^{(k)}}{2}\left(\|\textbf{s}\|_2^2-2\left\langle \textbf{s},\frac{\rho^{(k)}\textbf{w}^{(k+1)}-\boldsymbol{\gamma}^{(k)}_2}{\rho^{(k)}}\right\rangle\right)}\\
&=\argmin_{\textbf{s}\in\mathcal{W}_r}{\frac{\rho^{(k)}}{2}\left(\left\|\textbf{s}-\frac{\rho^{(k)}\textbf{w}^{(k+1)}-\boldsymbol{\gamma}^{(k)}_2}{\rho^{(k)}}\right\|_2^2\right)}\\
& =\Pi\left(\frac{\rho^{(k)}\textbf{w}^{(k+1)}-\boldsymbol{\gamma}^{(k)}_2}{\rho^{(k)}}\right)_{\mathcal{W}_r}
\end{split}    
\end{equation}

\end{document}